\documentclass[preprint,12pt]{elsarticle}
\usepackage{graphicx}
\usepackage{booktabs}
\usepackage{amsmath,amssymb}
\usepackage{multirow}
\usepackage{algorithm}
\usepackage{algpseudocode}
\usepackage{hyperref}

\usepackage{algorithm}
\usepackage{algpseudocode}
\usepackage{subcaption}

\usepackage{booktabs} 
\usepackage{siunitx}  

\usepackage{pgfplots}
\pgfplotsset{compat=1.18}
\begin{document}

\begin{frontmatter}

\title{
Secure-by-Disguise: A Systematic Evaluation of Image Disguising for Confidential Medical Image Modeling
}

\author[umbc]{Jason Rojas}
\ead{jasonr2@umbc.edu}

\author[umbc]{Jiajie He}
\ead{jiajieh1@umbc.edu}

\author[ltu]{Yash Patel}
\ead{ypatel3@ltu.edu}

\author[umbc]{Yuechun Gu}
\ead{ygu2@umbc.edu}

\author[uwm]{Zeyun Yu}
\ead{yuz@uwm.edu}

\author[umbc,cor1]{Keke Chen}
\ead{kekechen@umbc.edu}
\cortext[cor1]{Corresponding Author}

\affiliation[umbc]{organization={University of Maryland, Baltimore County}, 
addressline={1000 Hilltop Circle}, 
            city={Baltimore},
            postcode={21250}, 
            state={MD},
            country={USA}}

\affiliation[ltu]{organization={Lawrence Technological University}, 
addressline={21000 West Ten Mile Road}, 
            city={Southfield},
            postcode={48075}, 
            state={MI},
            country={USA}}

\affiliation[uwm]{organization={University of Wisconsin, Milwaukee}, 
addressline={3203 N. Downer Avenue}, 
            city={Milwaukee},
            postcode={53211}, 
            state={WI},
            country={USA}}

\begin{abstract}

\textbf{Objective:}
Cloud-based deep learning enables large-scale medical image analysis but raises significant privacy concerns when sensitive patient images are outsourced for model development. Image disguising has recently emerged as a promising privacy-enhancing technology (PET) that transforms images into visually unintelligible representations while preserving information for downstream learning. However, its suitability for clinically relevant medical image analysis remains largely unknown.

\textbf{Methods:}
We established a unified evaluation framework to systematically assess representative image disguising methods for confidential medical image analysis. Two representative frameworks, DisguisedNets and NeuraCrypt, were evaluated on four public medical imaging datasets spanning both image classification and semantic segmentation. Predictive utility, computational efficiency, parameter sensitivity, and robustness against representative reconstruction attacks were evaluated under a common experimental protocol.

\textbf{Results:}
Image disguising exhibited fundamentally different behavior on image-level and pixel-level medical AI tasks. While representative methods preserved practical utility for medical image classification, they incurred substantially greater performance degradation for dense semantic segmentation. Among the evaluated approaches, Randomized Multidimensional Transformation (RMT) consistently achieved the best balance between predictive performance, computational efficiency, and resistance to reconstruction attacks, whereas AES-based disguising resulted in severe utility degradation. Regression-based reconstruction attacks that were previously shown to be partially effective on natural-image benchmarks became considerably less effective on realistic medical images.

\textbf{Conclusion:}
Image disguising is a promising PET for confidential medical AI, particularly for cloud-based medical image classification. However, current approaches inadequately preserve the fine-grained spatial information required for dense medical image segmentation. These findings provide the first systematic evidence regarding the capabilities and limitations of image disguising for medical AI and offer practical guidance for the development of next-generation privacy-enhancing technologies for medical image analysis.

\end{abstract}
\begin{keyword}

Medical image analysis \sep
Privacy-preserving machine learning \sep
Image disguising \sep
Cloud computing \sep
Medical AI \sep
Deep learning

\end{keyword}

\end{frontmatter}

\section{Introduction}

Deep learning (DL) has become a core technology in medical imaging \cite{Litjens17,Hosny18,Miotto18,esteva2019guide}, achieving remarkable performance in a broad range of clinical applications \cite{cai20medimage,topol2019high}, including disease screening, lesion detection, tumor segmentation, and longitudinal disease monitoring. The rapid growth of medical imaging datasets and increasingly large deep learning models has simultaneously driven healthcare institutions toward cloud-based computing infrastructures that provide scalable storage and high-performance computational resources. While cloud platforms facilitate the development and deployment of medical AI systems, they also create an urgent need \cite{kelly2019key} for confidential medical AI that protects sensitive patient data throughout the model development lifecycle \cite{wiens2019do}.

Medical images, including computed tomography (CT), magnetic resonance imaging (MRI), ultrasound, and digital pathology images, contain highly sensitive patient information that may reveal anatomical characteristics and diagnostic findings. Unauthorized disclosure may have serious legal, ethical, and financial consequences, motivating regulations such as the General Data Protection Regulation (GDPR) \cite{gdpr} and the Health Insurance Portability and Accountability Act (HIPAA) \cite{hipaa}. Consequently, enabling collaborative or cloud-based medical image analysis while preserving patient confidentiality has become a fundamental challenge for trustworthy medical AI \cite{Kagadis13,knolle26,kaissis2020secure}.

A variety of privacy-enhancing technologies (PETs) \cite{nistpets,enisa2022pets} have been proposed to address this challenge. Cryptographic approaches, including homomorphic encryption (HE) \cite{chillotti20} and secure multiparty computation (SMC) \cite{mugunthan19}, provide strong confidentiality guarantees but typically incur computational costs several orders of magnitude higher than conventional deep learning. Data anonymization and de-identification techniques \cite{fung10} remove explicit identifiers but remain vulnerable to re-identification attacks using auxiliary information \cite{eliazar25}. Differential privacy (DP) \cite{dwork06,he25recps} provides rigorous mathematical privacy guarantees by perturbing the training process, but often sacrifices predictive performance \cite{ficek21}. As a result, existing PETs generally require a trade-off among privacy, model utility, and computational efficiency.

Among these approaches, \emph{image disguising} has recently emerged as an attractive alternative for confidential medical AI. Rather than encrypting computation or perturbing model optimization, image disguising transforms images into visually unintelligible representations while preserving sufficient information for downstream learning using conventional deep learning pipelines. Representative methods include DisguisedNets \cite{chen23toit}, which employs block-wise Randomized Multidimensional Transformation (RMT) or AES-based transformations, and NeuraCrypt \cite{yala21}, which generates confidential image representations for Vision Transformer (ViT) models \cite{vit21}. Compared with cryptographic methods, these approaches require only lightweight preprocessing while maintaining compatibility with existing training frameworks, making them attractive candidates for cloud-based medical AI.

However, whether image disguising is suitable for realistic medical image analysis remains largely unknown. Existing studies have been conducted almost exclusively on low-resolution natural-image classification benchmarks such as MNIST and CIFAR-10. In contrast, medical imaging often relies on subtle anatomical structures, fine-grained textures, and precise spatial relationships \cite{Litjens17}. Moreover, many clinically important applications, including lesion delineation and organ segmentation \cite{hesamian2019deep,minaee2021image}, require dense pixel-level prediction rather than image-level classification. These fundamental differences raise an important question: \emph{do conclusions regarding the utility and privacy of image disguising derived from natural-image benchmarks generalize to clinically relevant medical AI tasks?} Answering this question is essential for understanding the practical applicability and limitations of image disguising in confidential medical AI.

To address this question, we establish a unified evaluation framework for systematically assessing representative image disguising methods in confidential medical image analysis. Using this framework, we examine DisguisedNets and NeuraCrypt on four publicly available medical imaging datasets covering both image classification and semantic segmentation. Beyond predictive performance, we investigate computational efficiency, parameter sensitivity, and robustness against representative reconstruction attacks under a unified experimental protocol. To facilitate reproducible research, we further release the complete evaluation framework and experimental code as an open-source repository.

Our study yields three principal findings. First, current image disguising methods exhibit fundamentally different behavior on image-level and pixel-level clinical tasks, preserving substantially more utility for classification than for dense semantic segmentation. Second, among the evaluated methods, RMT consistently provides the best balance between predictive performance, computational efficiency, and resistance to reconstruction attacks, whereas AES-based disguising often causes severe utility degradation. Third, regression-based reconstruction attacks previously shown to be partially effective on low-resolution natural-image datasets become considerably less effective on realistic medical images. Together, these findings provide the first systematic evidence regarding the capabilities and limitations of image disguising for confidential medical AI and offer practical guidance for the development of next-generation privacy-enhancing technologies for medical image analysis.

The main contributions of this work are summarized as follows.

\begin{itemize}

\item We establish the first unified evaluation framework for assessing representative image disguising methods across clinically relevant medical image classification and semantic segmentation tasks.

\item Using this framework, we systematically characterize predictive utility, computational efficiency, parameter sensitivity, and robustness against reconstruction attacks, providing a comprehensive understanding of the practical utility-privacy trade-offs of image disguising.

\item We demonstrate that current image-disguising techniques exhibit fundamentally different behavior in image classification and dense semantic segmentation, providing practical guidance for future confidential medical AI systems.
\end{itemize}

The remainder of this paper is organized as follows. Section~\ref{sec:related} reviews related work. Section~\ref{sec:method} presents the evaluated image disguising methods and experimental setup. Section~\ref{sec:results} reports the experimental results. Section~\ref{sec:discussion} discusses the implications of the findings, and Section~\ref{sec:conclusion} concludes the paper.

\section{Related Work}
\label{sec:related}

\subsection{Privacy Enhancing Technologies for Outsourcing}
Privacy-preserving medical image analysis has attracted increasing attention with the growing adoption of cloud-based medical AI. Existing approaches can be broadly categorized into cryptographic methods, statistical privacy mechanisms, and data transformation techniques. Cryptographic approaches, including homomorphic encryption (HE) \cite{chillotti20} and secure multi-party computation (SMPC) \cite{mugunthan19}, provide strong confidentiality guarantees but incur substantial computational overhead that limits their scalability. Methods have been used in model inference for small models, e.g., CryptoNets \cite{gilad16} and SecureML\cite{mohassel17}, but they are not applicable to expensive outsourced training.  Differential privacy (DP) \cite{dwork06,he25recps} protects training data \cite{abadi16} by injecting carefully calibrated noise during optimization, often at the expense of prediction quality \cite{ficek21}. Record-level or local DP \cite{wang20ldp} also enables outsourced training. However, it results in greater utility loss. Conventional anonymization and de-identification techniques \cite{fung10} remove explicit identifiers but remain vulnerable to linkage and re-identification attacks \cite{eliazar25}. Compared with these approaches, image disguising transforms images into visually unintelligible representations while allowing standard deep learning models to operate directly on the transformed data, providing a computationally efficient alternative for confidential medical AI.

Federated learning (FL) has also emerged as an important paradigm for privacy-preserving medical AI by enabling multiple healthcare institutions to collaboratively train models without directly sharing patient data \cite{yang2019federated,sheller2020federated,ryan2024survey}. FL has demonstrated promising performance for medical image classification and segmentation across distributed clinical datasets. However, FL addresses a different privacy problem from image disguising. Instead of protecting data outsourced to an untrusted cloud, FL assumes that data remain locally stored and only model updates are exchanged. Moreover, recent studies have shown that model updates may still leak sensitive information through gradient inversion \cite{zhu19nips,geiping2020inverting}, membership inference \cite{hu22}, and reconstruction attacks \cite{haim22}, often requiring additional privacy mechanisms such as differential privacy or secure aggregation. Consequently, FL and image disguising should be viewed as complementary privacy-enhancing technologies targeting different deployment scenarios.

Most privacy-preserving medical imaging studies have focused on image classification tasks, including disease diagnosis and screening \cite{patricio23,Selvakumar25,kamal25,wang24ppimg}. In contrast, semantic segmentation has received comparatively less attention despite its importance in clinical applications \cite{gao25,skorupko25,bian21} such as tumor delineation, organ segmentation, and wound assessment. Unlike image classification, segmentation requires precise preservation of local spatial relationships \cite{simpson2019msd}, making it a particularly challenging setting for image disguising methods. To the best of our knowledge, no prior work has systematically compared the impact of image disguising on both medical image classification and semantic segmentation.

\subsection{Image Disguising and Security Evaluation}
Recent image disguising methods aim to conceal image content while preserving downstream learning utility. DisguisedNets \cite{chen23toit} combines block-wise permutation with Randomized Multidimensional Transformation (RMT) or AES-based transformations and demonstrates encouraging results on natural-image benchmarks such as MNIST and CIFAR-10. Unlike DisguisedNets, NeuraCrypt \cite{yala21} represents confidential representation learning rather than pixel-space transformation, enabling comparison between two fundamentally different design philosophies. Existing security analyses have primarily focused on ciphertext-only attacks and known-pair reconstruction attacks \cite{chen23toit,carlini21}, with evaluations conducted almost exclusively on low-resolution natural-image datasets.

However, whether these methods generalize to clinically realistic medical imaging remains largely unknown. Medical image analysis often depends on subtle anatomical structures and precise spatial relationships, particularly for dense prediction tasks such as semantic segmentation. To the best of our knowledge, no prior work has systematically evaluated representative image disguising methods across both medical image classification and semantic segmentation while jointly examining predictive utility, computational efficiency, and resistance to reconstruction attacks.  Unlike prior work that primarily proposed new disguising algorithms, this work focuses on establishing an evaluation methodology and deriving design principles for their applicability to realistic medical imaging tasks under a unified experimental protocol.

\section{Methods}
\label{sec:method}

\subsection{Unified Evaluation Framework}

We consider a cloud-based medical image analysis scenario in which a healthcare institution outsources deep learning training and inference to an untrusted cloud platform. Before outsourcing, each medical image is transformed locally using an image disguising function

\[
\tilde{x}_i=T(x_i;k),
\]

where $T(\cdot)$ is parameterized by a secret key $k$ known only to the data owner. The cloud therefore performs learning using the transformed dataset

\[
\tilde{\mathcal D}
=
\{
(\tilde{x}_i,\tilde y_i)
\}_{i=1}^{N}.
\]

For image classification, class labels remain unchanged
($\tilde y_i=y_i$).
For semantic segmentation, whenever the disguising method changes image geometry (e.g., block permutation), the same transformation is applied to the corresponding segmentation masks,

\[
\tilde y_i=T_y(y_i;k),
\]

to preserve pixel-wise correspondence.

We study their suitability for medical imaging applications, specifically, on three aspects:

\begin{itemize}
\item \textbf{Utility:} predictive performance on medical image classification and semantic segmentation.
\item \textbf{Privacy:} resistance to representative image reconstruction attacks.
\item \textbf{Efficiency:} computational overhead introduced by image disguising.
\end{itemize}

\begin{figure}[hbt!]
    \centering
    \includegraphics[width=0.9\textwidth]{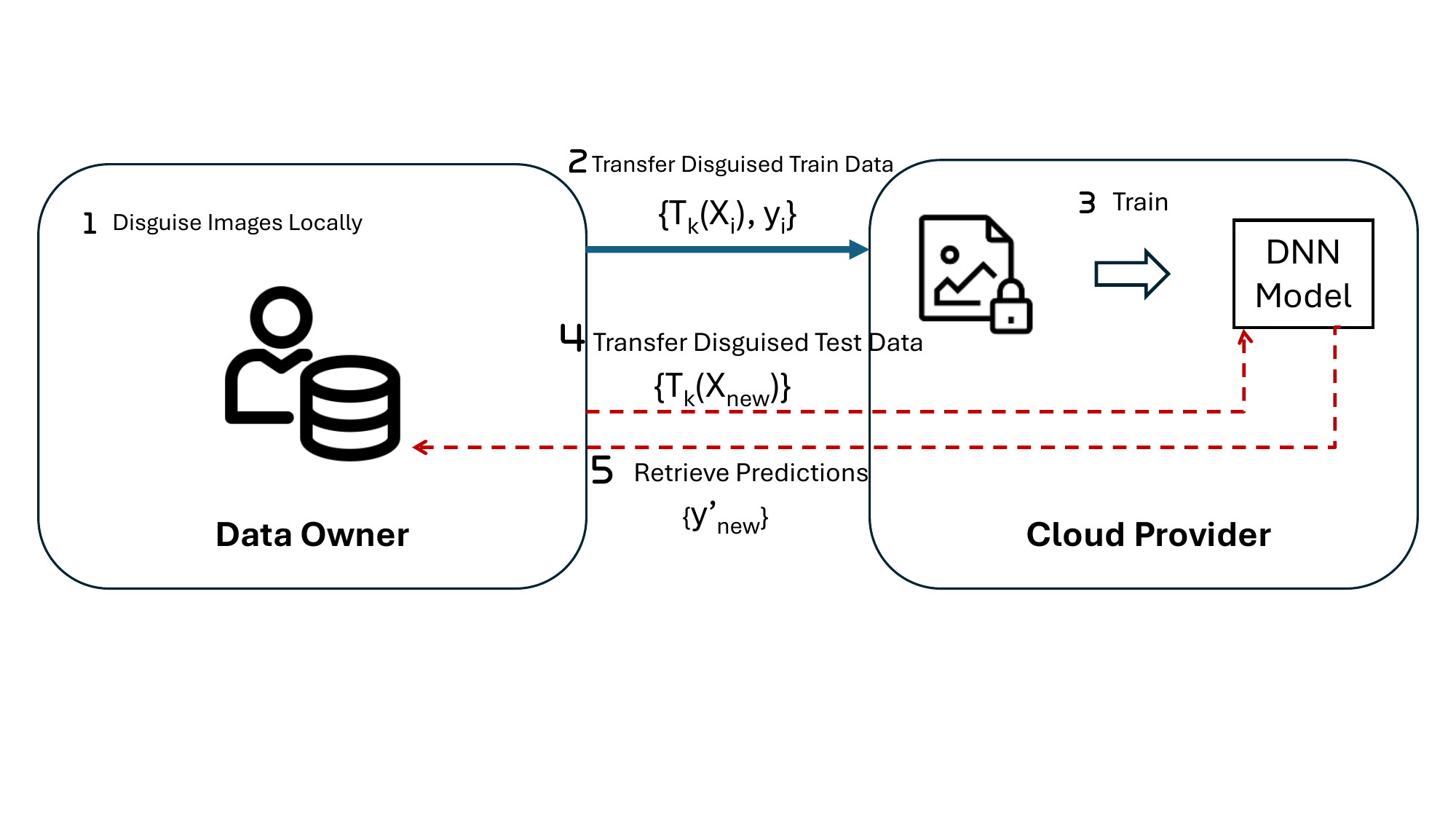}
    \caption{Image disguising framework for outsourced medical image analysis. Medical images are transformed locally before being uploaded to the cloud, allowing model training and inference without exposing the original patient images.}
    \label{fig:arch}
\end{figure}

As illustrated in Figure~\ref{fig:arch}, image disguising supports both outsourced model training and outsourced inference without exposing the original medical images to the cloud.

\subsection{Threat Model}

We adopt the threat model assumed in DisguisedNets. The cloud provider is considered \emph{honest-but-curious}: it correctly performs model training and inference but may attempt to infer sensitive patient information from the disguised images, training process, or trained model.

The primary assets are the original medical images and the secret transformation key. We consider two adversarial settings.

\begin{itemize}

\item \textbf{Weak adversary.}
The adversary knows the application domain, model architecture, and disguising algorithm, but not the transformation key.

\item \textbf{Strong adversary.}
The adversary additionally possesses a limited number of paired original and disguised images obtained through external leakage and attempts to reconstruct previously unseen images.

\end{itemize}

Accordingly, we evaluate representative reconstruction attacks and measure information leakage using pretrained DNN examiners. Poisoning attacks, adversarial examples, secure-channel attacks, and client compromise are beyond the scope of this work.

\subsection{Evaluated Image Disguising Methods}
We focus on two representative image disguising frameworks.

\textbf{DisguisedNets} \cite{chen23toit}
transforms images through
(1) blocktization and optional block-level permutation, followed by
(2) block-level transformation, which contains two transformation strategies: RMT and AES.
RMT performs an orthogonal transformation independently on each image block, optionally followed by additive random noise. AES instead encrypts each image block independently using standard cryptographic primitives. Both transformations preserve the image dimensions, allowing existing CNN architectures to be trained without modification. 

\begin{figure}[h]
\centering
\includegraphics[width=.9\linewidth]{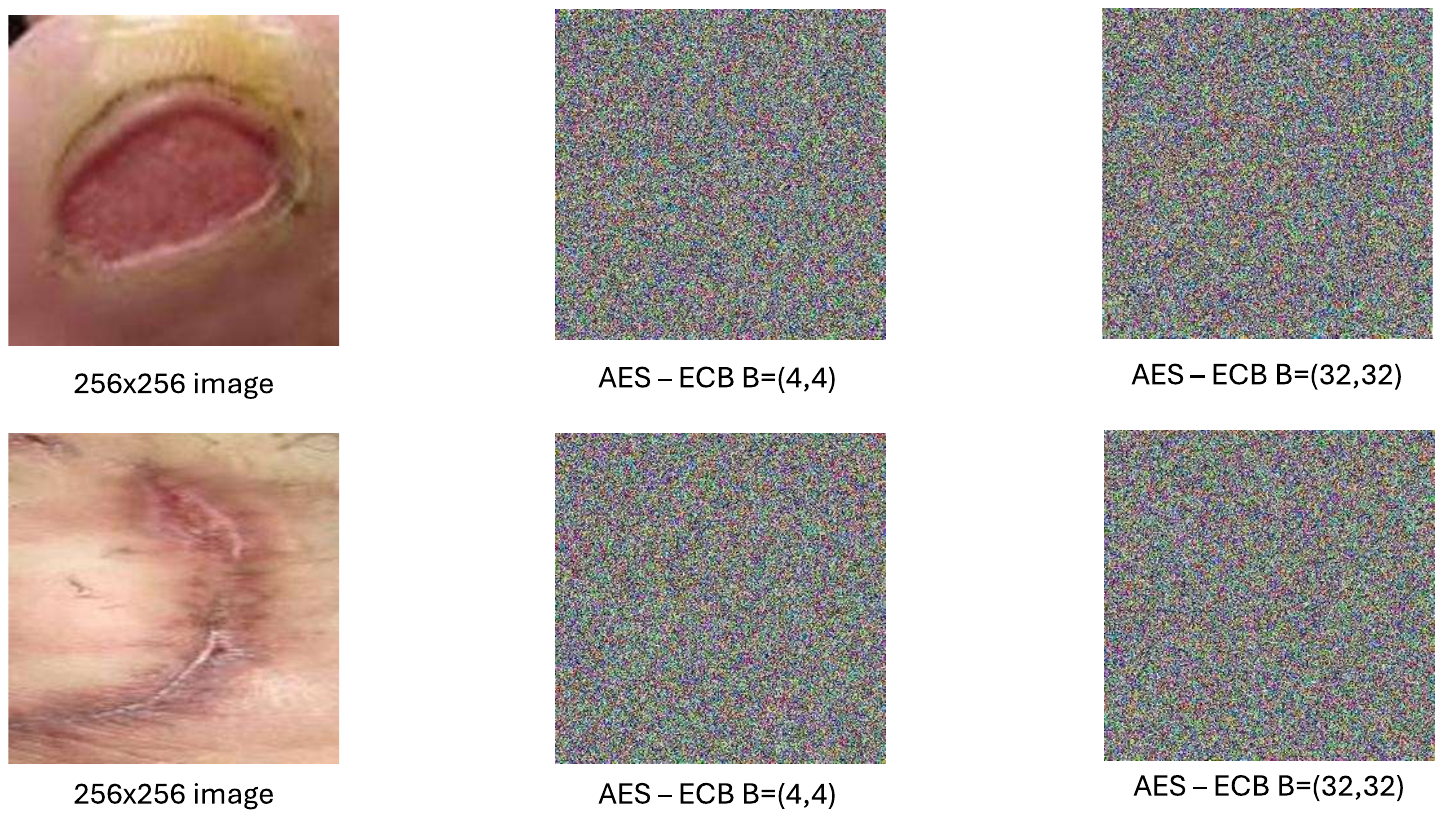}
\caption{Examples of AES-transformed images from the MData and Wound Patch datasets.}
\label{fig:aes_enc}
\end{figure}

\begin{itemize}

\item
\textbf{Randomized Multidimensional Transformation (RMT).}
Each image block is transformed as

\[
G(X_{ij})=R_j(X_{ij}+\Delta_{ij}),
\]

where $X_{ij}$ is a block of $X_i$ at position $j$, $R_j$ is a secret random orthogonal matrix for the position $j$ and is shared by all images, and $\Delta_{ij}$ denotes optional injected noise. We examine the effects of multiple block sizes and noise levels on medical image disguising.

\begin{figure}[h]
\centering
\includegraphics[width=.9\linewidth]{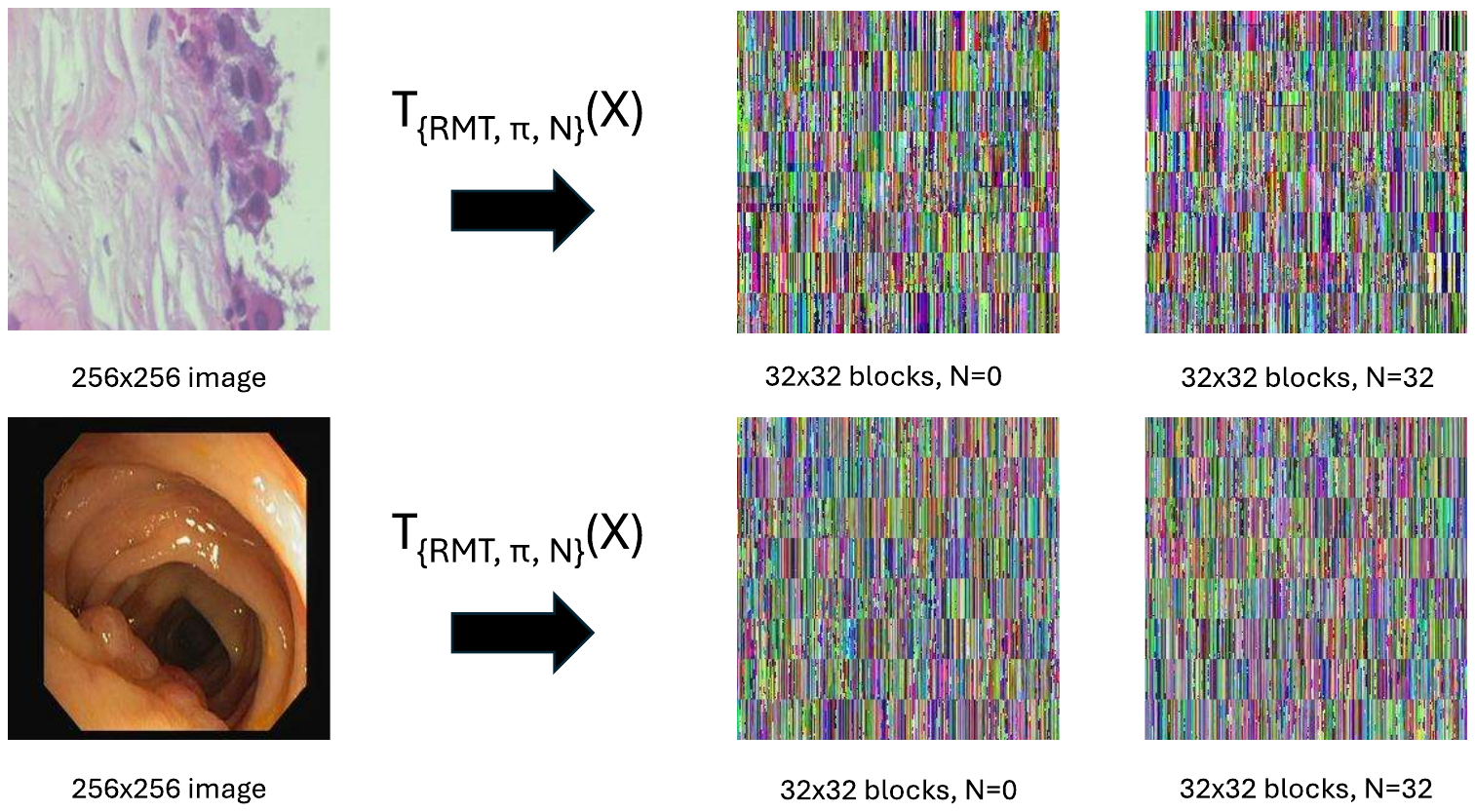}
\caption{Examples of block-wise RMT transformations on Breast and CVC datasets.}
\label{fig:disguises}
\end{figure}

\item
\textbf{AES Transformation.}
Each image block is encrypted independently using position-specific AES keys in ECB mode following the original implementation.
\end{itemize}
Representative transformed medical images are shown in Figures~\ref{fig:disguises} and \ref{fig:aes_enc}. Complete algorithmic details are available in the original DisguisedNets paper \cite{chen23toit}.

\textbf{NeuraCrypt} \cite{yala21}
uses a fundamentally different strategy by generating confidential patch embeddings through a private encoder before Vision Transformer learning.
We adopt the original implementation and parameter settings as a representative state-of-the-art baseline.
Unlike these pixel-space approaches, NeuraCrypt transforms image patches into confidential feature embeddings before downstream Vision Transformer (ViT) \cite{vit21} processing.

\subsection{Experimental Design}

\subsubsection{Datasets}
We select four publicly available medical imaging datasets covering both image classification and semantic segmentation tasks. The datasets span diverse imaging modalities, including histopathology, clinical wound photography, and colonoscopy, providing a comprehensive testbed for assessing whether image disguising methods developed on natural-image benchmarks generalize to realistic medical imaging workloads. Table~\ref{tab:datasets} summarizes the datasets used in this study.

\begin{table}[t]
\centering
\footnotesize
\caption{Summary of the datasets used in the evaluation.}
\label{tab:datasets}
\begin{tabular}{l l l c c l}
\toprule
\textbf{Dataset} & \textbf{Task} & \textbf{Modality} & \textbf{Images} & \textbf{Resolution} & \textbf{Output} \\
\midrule
Breast (PEM) & Classification & Histopathology & 1,820 & $700\times460$ & 2 classes \\
MData & Classification & Clinical wound & 738 & Variable & 6 classes \\
CVC-ClinicDB & Segmentation & Colonoscopy & 612 & $384\times288$ & Binary mask \\
Wound Patch & Segmentation & DFU photography & 110 & Variable & Binary mask \\
\bottomrule
\end{tabular}
\end{table}

Briefly, Breast and MData evaluate image classification under different imaging conditions, whereas CVC-ClinicDB and Wound Patch evaluate dense semantic segmentation requiring accurate preservation of spatial structures.

\subsubsection{Evaluated Models}
For DisguisedNets, we consider representative convolutional neural networks (CNNs), including VGG16, VGG19, ResNet34, and ResNet50 for image classification, and U-Net and UNet++ for semantic segmentation. Unless otherwise specified, subsequent ablation studies use ResNet34 and UNet++, which provide representative performance with substantially lower computational cost. NeuraCrypt is evaluated using the original Vision Transformer (ViT) architecture and implementation released by the authors.

\subsubsection{Experimental Protocol}
Unless otherwise specified, all images are resized to $256\times256$ before applying the disguising transformation. For DisguisedNets, we evaluate both RMT and AES-based transformations using multiple block sizes and parameter settings. NeuraCrypt follows the original implementation and recommended hyperparameters provided by the authors.

All CNN models are initialized with ImageNet-pretrained weights and fine-tuned end-to-end. Training uses the Adam optimizer with a learning rate of $10^{-4}$, weight decay of $10^{-4}$, batch size of 8, and 100 epochs. Image classification is optimized using cross-entropy loss, while semantic segmentation uses binary cross-entropy loss. Standard online data augmentation (random flipping, rotation, and normalization) was applied during training.

Performance is evaluated using stratified five-fold cross-validation. For each fold, three folds are used for training, one for validation, and one for testing. Identical data partitions are used for all compared methods to ensure fair comparison. Results are reported as the mean and 95\% confidence interval across the five folds.

All experiments are implemented in PyTorch and conducted on NVIDIA Ada A4000 GPUs. The complete experimental framework, including preprocessing, model training, evaluation, and attack implementations, is publicly available at \url{https://github.com/Jasonmix84/Secure-By-Disguise/}.

\subsubsection{Evaluation Metrics}
For image classification, predictive performance is measured using the macro F1-score. Semantic segmentation performance is evaluated using the Dice coefficient. Results are reported as the mean and 95\% confidence interval over the five cross-validation folds.

Following the threat model, privacy is evaluated through representative reconstruction attacks. Reconstruction quality is quantified using pretrained DNN examiners, where higher predictive performance on reconstructed images (e.g., F1 or Dice) indicates greater information leakage.

Computational efficiency is evaluated using image preprocessing time and model training cost, allowing comparison of the additional overhead introduced by different image disguising methods.

\section{Results}\label{sec:results}
Our experimental study focuses on three complementary perspectives: utility, efficiency, and privacy. We first examine whether current image disguising methods preserve the utility required for medical image classification and semantic segmentation across different deep learning architectures. We then quantify the computational overhead introduced by the disguising transformations. Finally, we investigate the influence of key transformation parameters through ablation studies and evaluate the resilience of the methods against representative reconstruction attacks for the most promising method RMT. Together, these experiments provide a comprehensive assessment of the practical applicability of image disguising for confidential medical image modeling.

\subsection{Utility Preservation for Image Classification}
We first compare the utility preservation of different image disguising methods for medical image classification.
Since DisguisedNets supports multiple transformation configurations, we use two representative settings throughout this comparison: a small block size ($4\times4$) and a large block size ($32\times32$), both without additional noise. The effects of block size and noise are investigated separately in the ablation studies. For NeuraCrypt, we use the original Vision Transformer (ViT) implementation and recommended parameter settings.

Figure~\ref{fig:breast-classification} summarizes the results on the Breast dataset. Models trained on the original images achieve consistently high performance across all CNN architectures, with macro-averaging F$_1$ scores ranging from 0.921 to 0.965. Among the evaluated CNNs, ResNet34 and ResNet50 slightly outperform the VGG models, while the ViT baseline achieves a lower F$_1$ score (0.844), indicating that conventional CNNs remain more effective for this histopathology classification task.

Across all CNN architectures, RMT consistently preserves substantially more utility than AES. The performance degradation introduced by RMT is relatively modest (approximately 10\% in F$_1$ score) and remains remarkably consistent across different network architectures, suggesting that its utility preservation is largely architecture-independent. In contrast, AES-based transformation leads to a dramatic reduction in classification performance, with F$_1$ scores approaching the level of random prediction. This observation indicates that the pixel representations produced by AES retain insufficient discriminative information for effective CNN learning. NeuraCrypt exhibits intermediate performance, outperforming AES while remaining below the utility achieved by RMT.

Figure~\ref{fig:mdata-classification} presents the corresponding results on the MData six-class wound classification dataset. Despite the substantially greater variation in image appearance, illumination, viewpoint, and background clutter, the overall trends remain consistent with those observed on the Breast dataset. RMT continues to provide the best balance between privacy and utility, AES now experiences less performance degradation, and NeuraCrypt still achieves intermediate performance.

Among the evaluated CNN architectures, ResNet34 consistently achieves performance comparable to ResNet50 while requiring substantially lower computational cost. Consequently, we adopt ResNet34 as the default classification model for the remaining experiments and ablation studies.

\definecolor{colorVGG16}{HTML}{1f77b4}   
\definecolor{colorVGG19}{HTML}{aec7e8}   
\definecolor{colorResNet34}{HTML}{ff7f0e} 
\definecolor{colorResNet50}{HTML}{ffbb78} 
\definecolor{colorViT}{HTML}{2ca02c}      

\begin{figure*}[htbp]
    \centering
    
    \centering
    \ref{lbl:VGG16} VGG16 \qquad
    \ref{lbl:VGG19} VGG19 \qquad
    \ref{lbl:ResNet34} ResNet34 \qquad
    \ref{lbl:ResNet50} ResNet50 \qquad
    \ref{lbl:ViT} ViT

    \begin{subfigure}[b]{\textwidth}
        \centering
        \begin{tikzpicture}
        \begin{axis} [
            ybar,
            enlarge x limits=0.25,
            ylabel={Mean $F_1$ Score},
            symbolic x coords={Original, RMT, AES, NeuraCrypt},
            xtick=data,
            ymin=0.4, ymax=1.05, 
            bar width=6pt,
            width=\textwidth,
            height=6.2cm,
            error bars/y dir=both,
            error bars/y explicit,
            error bars/error bar style={line width=0.6pt, draw=black!70},
            error bars/error mark options={rotate=90, mark size=1.5pt, line width=0.6pt}
        ]
        
        \addplot[fill=colorVGG16, draw=black!80] coordinates {
            (Original, 0.9405) +- (0, 0.0214) 
            (RMT, 0.7949) +- (0, 0.0236)
            (AES, 0.5000) +- (0, 0.0000)
            (NeuraCrypt, 0) +- (0, 0)
        }; \label{lbl:VGG16} 
        
        \addplot[fill=colorVGG19, draw=black!80] coordinates {
            (Original, 0.9211) +- (0, 0.0115) 
            (RMT, 0.7961) +- (0, 0.0247)
            (AES, 0.5000) +- (0, 0.0000)
            (NeuraCrypt, 0) +- (0, 0)
        }; \label{lbl:VGG19} 
        
        \addplot[fill=colorResNet34, draw=black!80] coordinates {
            (Original, 0.9636) +- (0, 0.0116) 
            (RMT, 0.8268) +- (0, 0.0292)
            (AES, 0.5057) +- (0, 0.0205)
            (NeuraCrypt, 0) +- (0, 0)
        }; \label{lbl:ResNet34} 
        
        \addplot[fill=colorResNet50, draw=black!80] coordinates {
            (Original, 0.9647) +- (0, 0.0045) 
            (RMT, 0.8271) +- (0, 0.0150)
            (AES, 0.5002) +- (0, 0.0009)
            (NeuraCrypt, 0) +- (0, 0)
        }; \label{lbl:ResNet50} 
        
        \addplot[fill=colorViT, draw=black!80] coordinates {
            (Original, 0.8437) +- (0, 0.0170) 
            (RMT, 0) +- (0, 0) 
            (AES, 0) +- (0, 0) 
            (NeuraCrypt, 0.7070) +- (0, 0.0238)
        }; \label{lbl:ViT} 
        \end{axis}
        \end{tikzpicture}
        \caption{Breast Classification Performance}
        \label{fig:breast-classification}
    \end{subfigure}
    \hfill 
    \begin{subfigure}[b]{\textwidth}
        \centering
        \begin{tikzpicture}
        \begin{axis} [
            ybar,
            enlarge x limits=0.25,
            ylabel={Mean $F_1$ Score},
            symbolic x coords={Original, RMT, AES, NeuraCrypt},
            xtick=data,
            ymin=0.1, ymax=0.85, 
            bar width=6pt,
            width=\textwidth,
            height=6.2cm,
            error bars/y dir=both,
            error bars/y explicit,
            error bars/error bar style={line width=0.6pt, draw=black!70},
            error bars/error mark options={rotate=90, mark size=1.5pt, line width=0.6pt}
        ]
        
        \addplot[fill=colorVGG16, draw=black!80] coordinates {
            (Original, 0.7379) +- (0, 0.0323) 
            (RMT, 0.5612) +- (0, 0.0492)
            (AES, 0.3319) +- (0, 0.0460)
            (NeuraCrypt, 0) +- (0, 0)
        };
        
        \addplot[fill=colorVGG19, draw=black!80] coordinates {
            (Original, 0.7198) +- (0, 0.0371) 
            (RMT, 0.5498) +- (0, 0.0599)
            (AES, 0.3294) +- (0, 0.0528)
            (NeuraCrypt, 0) +- (0, 0)
        };
        
        \addplot[fill=colorResNet34, draw=black!80] coordinates {
            (Original, 0.7537) +- (0, 0.0243) 
            (RMT, 0.5854) +- (0, 0.0255)
            (AES, 0.3478) +- (0, 0.0302)
            (NeuraCrypt, 0) +- (0, 0)
        };
        
        \addplot[fill=colorResNet50, draw=black!80] coordinates {
            (Original, 0.7243) +- (0, 0.0344) 
            (RMT, 0.5837) +- (0, 0.0485)
            (AES, 0.3763) +- (0, 0.0410)
            (NeuraCrypt, 0) +- (0, 0)
        };
        
        \addplot[fill=colorViT, draw=black!80] coordinates {
            (Original, 0.6584) +- (0, 0.0536) 
            (RMT, 0) +- (0, 0) 
            (AES, 0) +- (0, 0) 
            (NeuraCrypt, 0.3937) +- (0, 0.0537)
        };
        \end{axis}
        \end{tikzpicture}
        \caption{MData Classification Performance}
        \label{fig:mdata-classification}
    \end{subfigure}

    \caption{Comparison of Mean $F_1$ scores across different methods grouped by model architecture for both datasets. Error bars denote confidence intervals.}
    \label{fig:combined-classification-results}
\end{figure*}
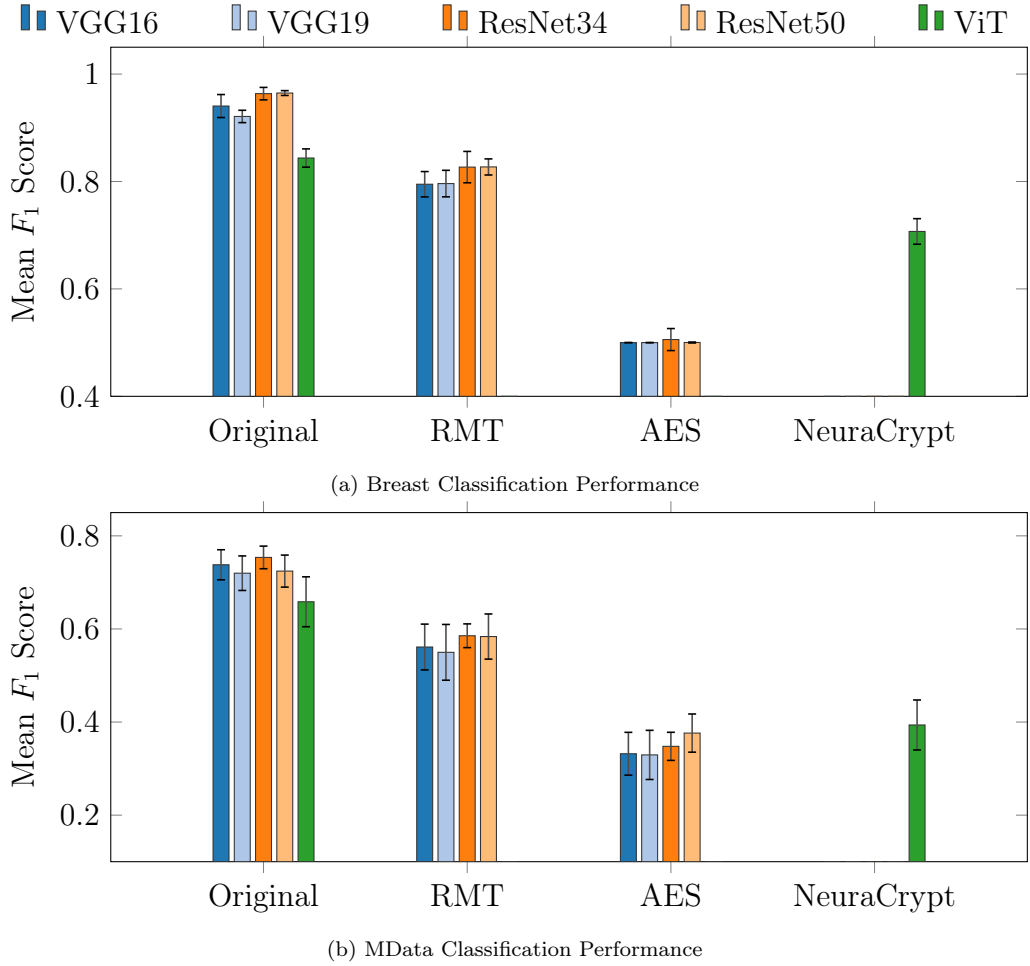

\subsection{Utility Preservation for Segmentation Tasks}
The empirical results from the CVC and Woundpatch segmentation experiments reveal several important characteristics of image disguising methods for the segmentation task. We have evaluated with the popular metrics: Dice and IoU \cite{milletari2016v}. Since they are closely related, we only show the Dice results.

In the untreated reference configurations (Original), both UNet and UNet++ achieve elite, nearly identical performance with mean Dice scores of approximately 0.92 for CVC and 0.87 for Wound Patch. Conversely, the Vision Transformer (ViT - Original) experiences a severe structural breakdown across both tasks, much worse than CNN-based architectures. 

RMT results in uniform, stable performance degradation across both convolutional frameworks. For the CVC dataset, the aggregated Mean Dice score for UNet and UNet++ settles almost identically at $\sim0.48$, while on the Woundpatch dataset, both architectures stabilize in a tight band of $\sim0.68$ to $0.69$. RMT acts in an architecture-agnostic fashion, processing features uniformly regardless of whether a standard UNet or a nested UNet++ block structure is deployed. Because RMT preserves predictable mid-tier utility while allowing the model to steadily minimize its loss function, these shared networks successfully capture residual structural variations.

The integration of AES transformed data exposes a deep dependency on the underlying dataset domain. On the CVC dataset, the highly non-linear pixel diffusion of AES triggers a catastrophic structural blackout for convolutional models, forcing UNet and UNet++ down to virtually unlearnable segmentation boundaries (Mean Dice of $\sim0.24$ to $0.25$).  However, on the Woundpatch dataset, the models on AES, maintaining a moderate utility band of approximately 0.62 to 0.63 Mean Dice. This divergence shows the AES encrypted images strictly dependent on the baseline structural layout and geometric complexity of the source imaging domain. 

Finally, although the native ViT baseline is already severely compromised on raw data, NeuraCrypt does not compromise the performance further. The Dice values remain almost unchanged in both datasets.

\definecolor{colorUNet}{HTML}{1f77b4}    
\definecolor{colorUNetPP}{HTML}{aec7e8}  
\definecolor{colorViT}{HTML}{2ca02c}     

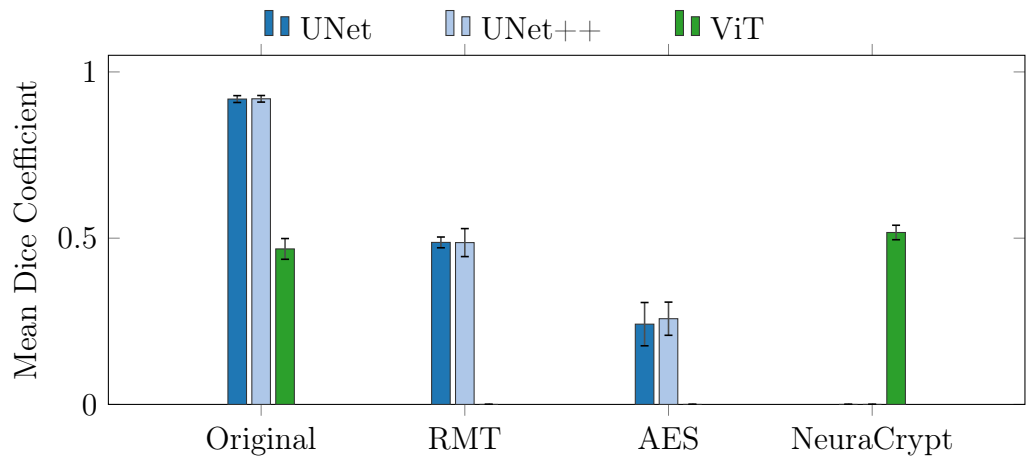
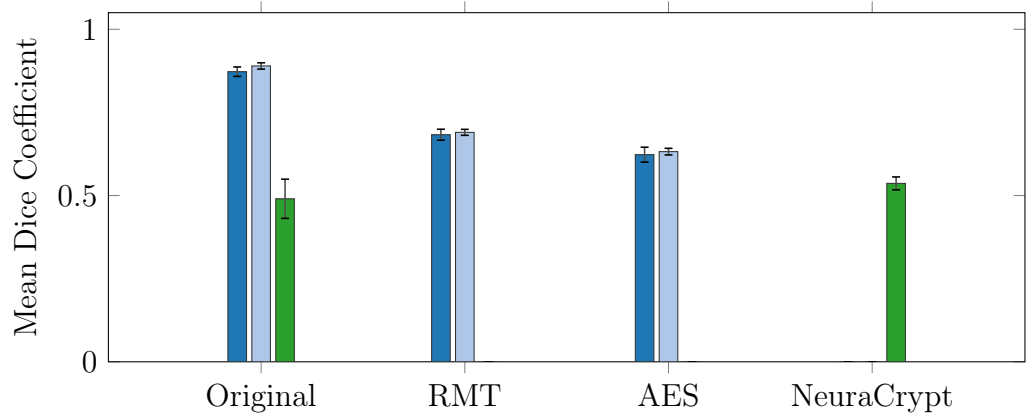
\begin{figure*}[htbp]
    \centering
    
    \centering
    \ref{lbl:UNet} UNet \qquad
    \ref{lbl:UNetPP} UNet++ \qquad
    \ref{lbl:SegViT} ViT

    \begin{subfigure}[b]{\textwidth}
        \centering
        \begin{tikzpicture}
        \begin{axis} [
            ybar,
            enlarge x limits=0.25,
            ylabel={Mean Dice Coefficient},
            symbolic x coords={Original, RMT, AES, NeuraCrypt},
            xtick=data,
            ymin=0.0, ymax=1.05, 
            bar width=7pt,
            width=\textwidth,
            height=6.2cm,
            error bars/y dir=both,
            error bars/y explicit,
            error bars/error bar style={line width=0.6pt, draw=black!70},
            error bars/error mark options={rotate=90, mark size=1.5pt, line width=0.6pt}
        ]
        
        \addplot[fill=colorUNet, draw=black!80] coordinates {
            (Original, 0.9184) +- (0, 0.0103) 
            (RMT, 0.4873) +- (0, 0.0162)
            (AES, 0.2414) +- (0, 0.0651)
            (NeuraCrypt, 0) +- (0, 0)
        }; \label{lbl:UNet} 
        
        \addplot[fill=colorUNetPP, draw=black!80] coordinates {
            (Original, 0.9192) +- (0, 0.0099) 
            (RMT, 0.4867) +- (0, 0.0421)
            (AES, 0.2578) +- (0, 0.0499)
            (NeuraCrypt, 0) +- (0, 0)
        }; \label{lbl:UNetPP} 
        
        \addplot[fill=colorViT, draw=black!80] coordinates {
            (Original, 0.4677) +- (0, 0.0312) 
            (RMT, 0) +- (0, 0) 
            (AES, 0) +- (0, 0) 
            (NeuraCrypt, 0.5171) +- (0, 0.0218)
        }; \label{lbl:SegViT} 
        \end{axis}
        \end{tikzpicture}
        \caption{CVC Dataset}
        \label{fig:cvc-dice}
    \end{subfigure}
    \hfill 
    \begin{subfigure}[b]{\textwidth}
        \centering
        \begin{tikzpicture}
        \begin{axis} [
            ybar,
            enlarge x limits=0.25,
            ylabel={Mean Dice Coefficient},
            symbolic x coords={Original, RMT, AES, NeuraCrypt},
            xtick=data,
            ymin=0.0, ymax=1.05, 
            bar width=7pt,
            width=\textwidth,
            height=6.2cm,
            error bars/y dir=both,
            error bars/y explicit,
            error bars/error bar style={line width=0.6pt, draw=black!70},
            error bars/error mark options={rotate=90, mark size=1.5pt, line width=0.6pt}
        ]
        
        \addplot[fill=colorUNet, draw=black!80] coordinates {
            (Original, 0.8726) +- (0, 0.0142) 
            (RMT, 0.6830) +- (0, 0.0165)
            (AES, 0.6230) +- (0, 0.0224)
            (NeuraCrypt, 0) +- (0, 0)
        };
        
        \addplot[fill=colorUNetPP, draw=black!80] coordinates {
            (Original, 0.8898) +- (0, 0.0095) 
            (RMT, 0.6901) +- (0, 0.0090)
            (AES, 0.6321) +- (0, 0.0100)
            (NeuraCrypt, 0) +- (0, 0)
        };
        
        \addplot[fill=colorViT, draw=black!80] coordinates {
            (Original, 0.4903) +- (0, 0.0590) 
            (RMT, 0) +- (0, 0) 
            (AES, 0) +- (0, 0) 
            (NeuraCrypt, 0.5366) +- (0, 0.0194)
        };
        \end{axis}
        \end{tikzpicture}
        \caption{Woundpatch Dataset}
        \label{fig:woundpatch-dice}
    \end{subfigure}

    \caption{Comparison of Mean Dice coefficients across different privacy methods grouped by model architecture for CVC and Woundpatch datasets.}
    \label{fig:combined-segmentation-results}
\end{figure*}

\begin{figure*}[htbp]
    \centering
    
    \begin{subfigure}[b]{0.32\textwidth}
        \centering
        \includegraphics[width=\textwidth]{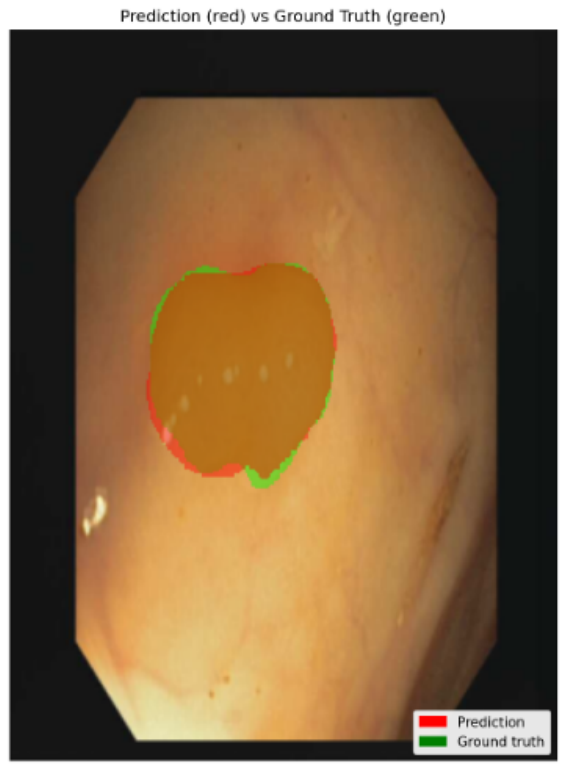}
        \caption{Original Data}
        \label{fig:cvc-org}
    \end{subfigure}
    \hfill 
    \begin{subfigure}[b]{0.32\textwidth}
        \centering
        \includegraphics[width=\textwidth]{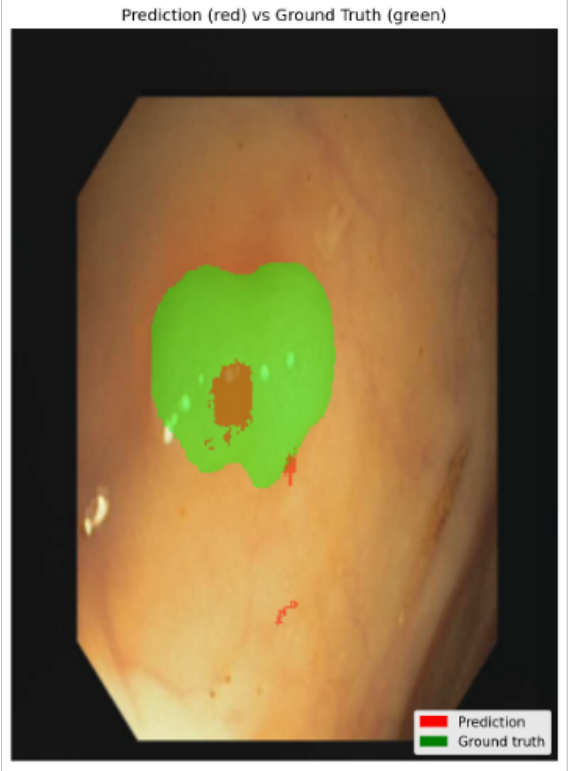}
        \caption{RMT Transformed}
        \label{fig:cvc-rmt}
    \end{subfigure}
    \hfill 
    \begin{subfigure}[b]{0.32\textwidth}
        \centering
        \includegraphics[width=\textwidth]{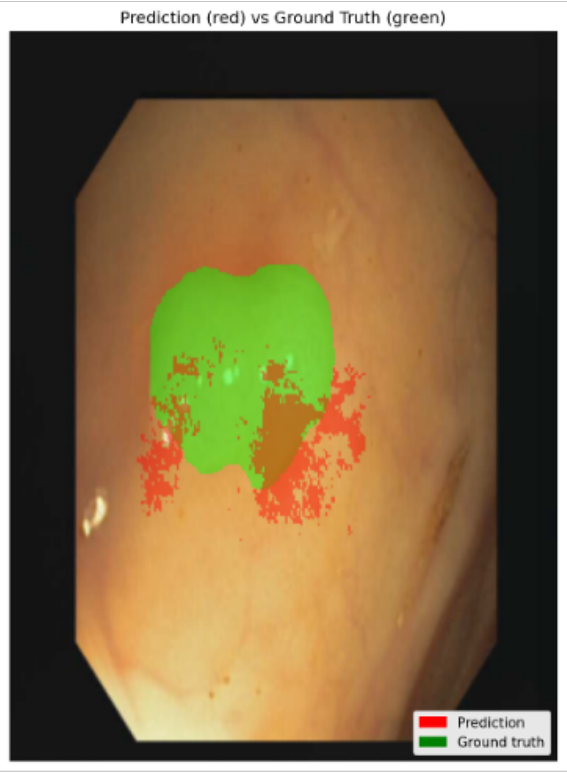}
        \caption{AES Transformed}
        \label{fig:cvc-aes}
    \end{subfigure}

    \caption{Visual comparison of dense semantic segmentation results on the CVC dataset under different image configurations: the polyp mask on a colonoscopy image. The green marks are the ground truth and the red are predicted. (a) baseline prediction using unmodified original images, showing almost perfect predction. (b) prediction using RMT disguised data, which gives only a small positive patch within the target area, and (c) prediction using AES encrypted data, which gives more detected pixels, but many are out of the target area.}
    \label{fig:cvc-segmentation-comparisons}
\end{figure*}

\begin{figure*}[htbp]
    \centering
    
    \begin{subfigure}[b]{0.32\textwidth}
        \centering
        \includegraphics[width=\textwidth]{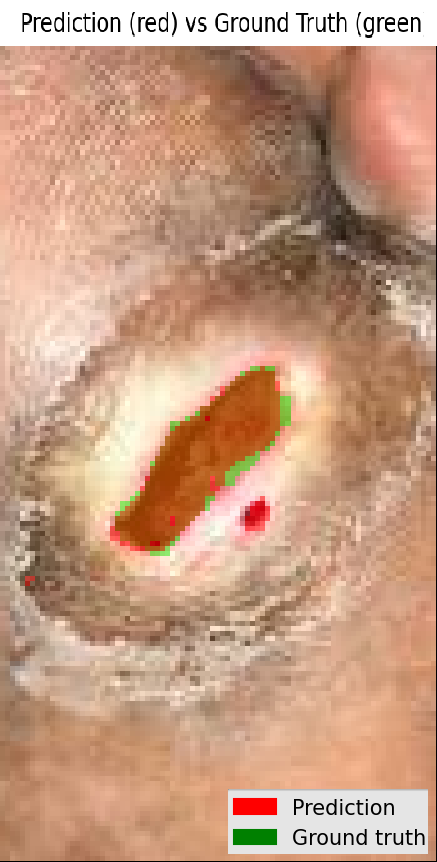}
        \caption{Original Data}
        \label{fig:wp-org}
    \end{subfigure}
    \hfill 
    \begin{subfigure}[b]{0.32\textwidth}
        \centering
        \includegraphics[width=\textwidth]{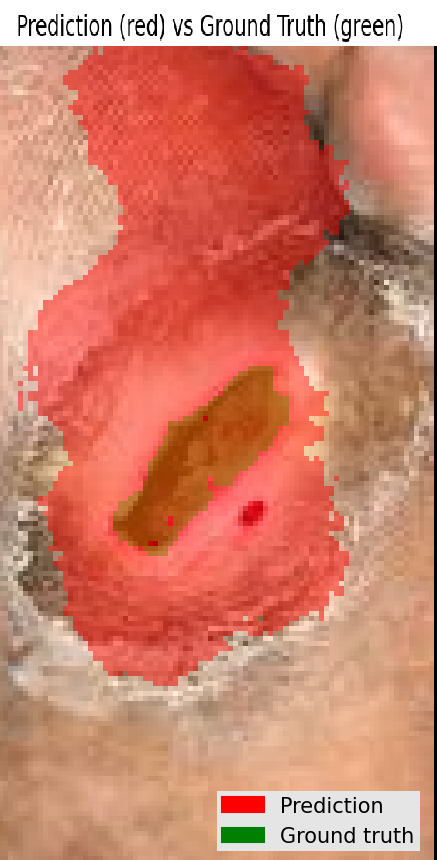}
        \caption{RMT Transformed}
        \label{fig:wp-rmt}
    \end{subfigure}
    \hfill 
    \begin{subfigure}[b]{0.32\textwidth}
        \centering
        \includegraphics[width=\textwidth]{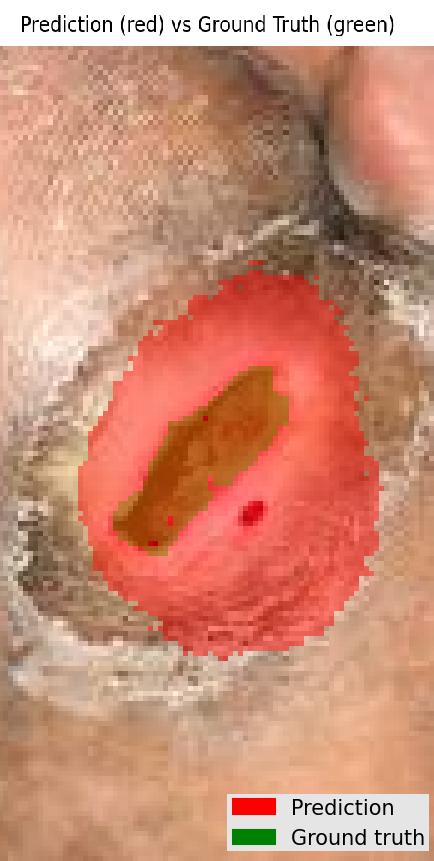}
        \caption{AES Transformed}
        \label{fig:wp-aes}
    \end{subfigure}

    \caption{Visual comparison of dense semantic segmentation results on the Wound-Patch dataset under different image configurations: the selected sample is a foot wound caused by diabetes. (a) Baseline prediction using unmodified original images, where the predicted area overlaps the ground truth area very well. (b) Prediction using RMT disguised data, which mistakenly predicts a large area as the wound, although the predicted area contains the ground truth area. (c) Prediction using AES encrypted data, which covers the target area and the errors are smaller than RMT's.}
    \label{fig:wp-segmentation-comparisons}
\end{figure*}

\subsection{Computational Efficiency}

One of the major advantages of image disguising over cryptographic approaches, such as homomorphic encryption and secure multi-party computation, is its low computational overhead. We therefore evaluate computational efficiency from two perspectives: (1) the preprocessing cost required to disguise images before training and (2) the influence of image disguising on model convergence during training.

Table~\ref{tab:runtime} summarizes the preprocessing time on the Breast (PEM) dataset containing 1,820 images. Both RMT and AES introduce relatively modest overhead, with average processing times ranging from a few milliseconds to several tens of milliseconds per image, making both methods practical for offline dataset preparation.

The preprocessing cost of RMT decreases substantially as the block size increases. With a small block size ($4\times4$), RMT requires 134.7 seconds (74.0 ms per image) because a large number of independent block transformations must be performed. Increasing the block size to $64\times64$ reduces the processing time to 6.8 seconds (3.7 ms per image). In comparison, AES preprocessing is less sensitive to block size, decreasing from 86.6 seconds (47.6 ms per image) to 49.0 seconds (26.9 ms per image). Across all evaluated settings, both methods incur only lightweight preprocessing overhead relative to the subsequent model training process.

\begin{table}[htbp]
    \centering
    \caption{Preprocessing time of image disguising methods on the Breast (PEM) dataset (1,820 images).}
    \label{tab:runtime}
    \begin{tabular}{@{} c c c c c @{}}
        \toprule
        & \multicolumn{2}{c}{\textbf{RMT}} & \multicolumn{2}{c}{\textbf{AES}} \\
        \cmidrule(lr){2-3} \cmidrule(lr){4-5}
        \textbf{Block Size} & \textbf{Total (s)} & \textbf{Per Image (ms)} & \textbf{Total (s)} & \textbf{Per Image (ms)} \\
        \midrule
        $4\times4$   & 134.7 & 74.01 & 86.6 & 47.58 \\
        $16\times16$ & 14.9  & 8.19  & 51.1 & 28.08 \\
        $64\times64$ & 6.8   & 3.74  & 49.0 & 26.92 \\
        \bottomrule
    \end{tabular}
\end{table}

Beyond preprocessing, we further examine whether image disguising affects the optimization process during model training. To illustrate the optimization behavior, we use the representative RMT configuration with block size 2x2. Similar trends were observed for other RMT configurations. Figure~\ref{fig:global_learning_curves} compares the learning curves obtained using the original images and RMT-disguised images. Both models converge at approximately the same epoch, reaching their lowest validation loss at Epoch~13 and Epoch~14, respectively. This observation indicates that image disguising does not substantially increase the number of training epochs required for convergence.

Nevertheless, the learning dynamics differ after convergence. Compared with training on the original images, models trained on RMT-disguised images exhibit a higher minimum validation loss and begin to overfit earlier, as evidenced by the increasing validation loss after the optimal epoch. These results suggest that although RMT introduces little additional optimization cost, the transformed images produce a more challenging learning problem and reduce the achievable validation performance. Consequently, standard training strategies such as early stopping become increasingly important when training on disguised medical images.

Overall, image disguising introduces only a lightweight computational overhead. The additional cost is dominated by a one-time preprocessing step, while the subsequent model training converges at a rate comparable to that of plaintext images. These findings support one of the primary motivations of image disguising: providing practical privacy protection for medical image modeling without the substantial computational burden associated with cryptographic learning methods.

\definecolor{trainColor}{HTML}{2980b9} 
\definecolor{valColor}{HTML}{c0392b}   

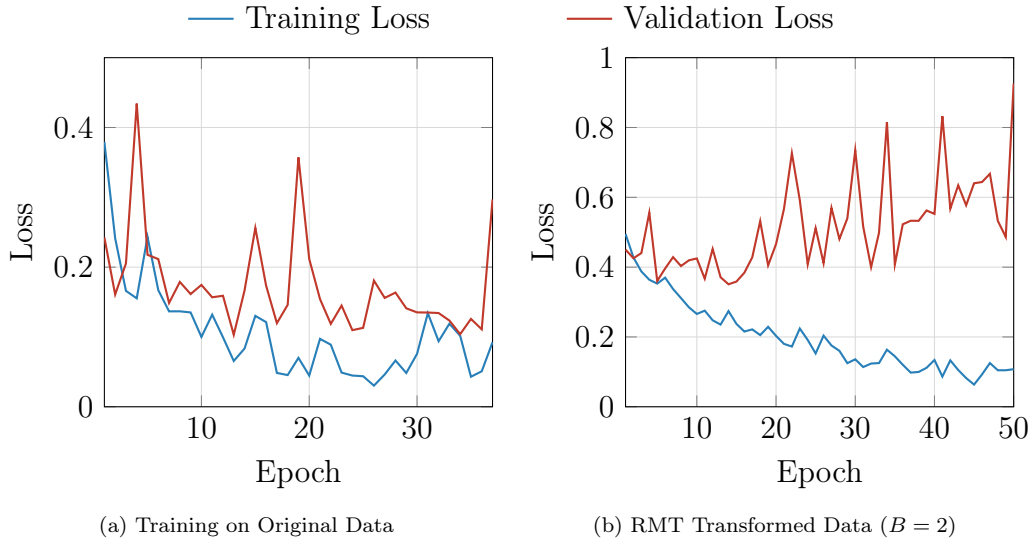
\begin{figure*}[htbp]
    \centering
    
    \ref{lbl:trainLoss} Training Loss \qquad\qquad \ref{lbl:valLoss} Validation Loss

    \begin{subfigure}[b]{0.49\textwidth}
        \centering
        \begin{tikzpicture}
        \begin{axis}[
            xlabel={Epoch},
            ylabel={Loss},
            xmin=1, xmax=37,
            ymin=0.0, ymax=0.50, 
            grid=both,
            grid style={line width=.1pt, draw=gray!10},
            major grid style={line width=.2pt, draw=gray!30},
            width=\textwidth,
            height=6.2cm,
        ]

        \addplot[color=trainColor, thick, mark=none] coordinates {
            (1,0.37931463) (2,0.2413848307) (3,0.1660422766) (4,0.155327948) 
            (5,0.2434365395) (6,0.1670056207) (7,0.1366638238) (8,0.1367181387) 
            (9,0.135105987) (10,0.1000238552) (11,0.1319456804) (12,0.09955816195) 
            (13,0.06555955864) (14,0.08340710238) (15,0.1302164527) (16,0.1212389604) 
            (17,0.04845610386) (18,0.04538908678) (19,0.07005896788) (20,0.04470182402) 
            (21,0.0972841332) (22,0.08888322486) (23,0.0489389785) (24,0.0447182402) 
            (25,0.0437326553) (26,0.0302657347) (27,0.0462656576) (28,0.0662688193) 
            (29,0.0483807581) (30,0.0758106134) (31,0.1340791457) (32,0.0938600784) 
            (33,0.1189689743) (34,0.101656186) (35,0.0430229223) (36,0.0508260492) 
            (37,0.0924124384)
        }; \label{lbl:trainLoss}

        \addplot[color=valColor, thick, mark=none] coordinates {
            (1,0.2424205829) (2,0.1608346099) (3,0.2051856543) (4,0.4342525264) 
            (5,0.2176621171) (6,0.2115106496) (7,0.1486810401) (8,0.1787788418) 
            (9,0.1614143982) (10,0.1745509547) (11,0.1567880034) (12,0.158999327) 
            (13,0.1031319523) (14,0.1666188977) (15,0.2561684247) (16,0.1735182714) 
            (17,0.1197229431) (18,0.1459779636) (19,0.3571705782) (20,0.2116449731) 
            (21,0.1541988894) (22,0.1186925196) (23,0.1449313017) (24,0.1097386404) 
            (25,0.1130258358) (26,0.1807140368) (27,0.1558934186) (28,0.1635560474) 
            (29,0.1409891801) (30,0.1351911928) (31,0.1349860172) (32,0.1342252878) 
            (33,0.1235094142) (34,0.1040048308) (35,0.1258209124) (36,0.1109086782) 
            (37,0.2969240079)
        }; \label{lbl:valLoss}

        \end{axis}
        \end{tikzpicture}
        \caption{Training on Original Data}
        \label{fig:loss_original}
    \end{subfigure}
    \hfill
    \begin{subfigure}[b]{0.49\textwidth}
        \centering
        \begin{tikzpicture}
        \begin{axis} [
            xlabel={Epoch},
            ylabel={Loss},
            xmin=1, xmax=50,
            ymin=0.0, ymax=1.00, 
            grid=both,
            grid style={line width=.1pt, draw=gray!10},
            major grid style={line width=.2pt, draw=gray!30},
            width=\textwidth,
            height=6.2cm,
        ]

        \addplot[color=trainColor, thick, mark=none] coordinates {
            (1,0.49498435) (2,0.4273865) (3,0.38760717) (4,0.36386011) (5,0.3527445) 
            (6,0.36980922) (7,0.33728661) (8,0.31135964) (9,0.28498042) (10,0.26579184) 
            (11,0.27529283) (12,0.24765328) (13,0.23547349) (14,0.2739253) (15,0.23739959) 
            (16,0.21555024) (17,0.22161414) (18,0.20529097) (19,0.2290366) (20,0.20310088) 
            (21,0.18015771) (22,0.17248222) (23,0.22413319) (24,0.19132054) (25,0.15273123) 
            (26,0.20376764) (27,0.17549868) (28,0.16014708) (29,0.1249165) (30,0.13598239) 
            (31,0.11378192) (32,0.12378854) (33,0.12524523) (34,0.1635399) (35,0.14459616) 
            (36,0.12032981) (37,0.09786698) (38,0.09971499) (39,0.11153677) (40,0.13397405) 
            (41,0.08652683) (42,0.13293646) (43,0.10527189) (44,0.08223094) (45,0.06355712) 
            (46,0.09259632) (47,0.12521115) (48,0.10454531) (49,0.10459778) (50,0.10762418)
        };

        \addplot[color=valColor, thick, mark=none] coordinates {
            (1,0.45016874) (2,0.42538259) (3,0.44038958) (4,0.55586613) (5,0.3603067) 
            (6,0.39616833) (7,0.42863027) (8,0.40315356) (9,0.4195227) (10,0.42488165) 
            (11,0.36690607) (12,0.45196092) (13,0.37095286) (14,0.350857) (15,0.358562) 
            (16,0.38393917) (17,0.42821899) (18,0.53170022) (19,0.40505161) (20,0.46595863) 
            (21,0.56659488) (22,0.72625106) (23,0.59177319) (24,0.40819496) (25,0.51177821) 
            (26,0.41150975) (27,0.56949163) (28,0.48071663) (29,0.53941972) (30,0.73447991) 
            (31,0.51633799) (32,0.39983727) (33,0.49847887) (34,0.81548482) (35,0.40758209) 
            (36,0.52273395) (37,0.53284868) (38,0.53260297) (39,0.56231731) (40,0.55243895) 
            (41,0.83223196) (42,0.56766148) (43,0.63429963) (44,0.57714661) (45,0.64019233) 
            (46,0.64372171) (47,0.66727917) (48,0.5324403) (49,0.48649498) (50,0.92607922)
        };

        \end{axis}
        \end{tikzpicture}
        \caption{RMT Transformed Data ($B=2$)}
        \label{fig:loss_rmt}
    \end{subfigure}

    \caption{Training and validation loss trajectories for the Breast classifier: panel (a) demonstrates standard convergence behavior on raw clinical imagery, while panel (b) showcases the challenging optimization terrain and rapid late-stage overfitting induced by the RMT framework.}
    \label{fig:global_learning_curves}
\end{figure*}

\subsection{Ablation Study of RMT}
Among the evaluated image disguising methods, RMT consistently achieves the best balance between utility and computational efficiency. We therefore conduct a detailed ablation study to understand how its key design parameters influence downstream performance. Specifically, we investigate the effects of block size and injected noise on both image classification and semantic segmentation, providing practical guidance for selecting RMT parameters in medical imaging applications. We also show that RMT is surprisingly resilient to attacks under the strong adversary setting. 

\subsubsection{Effect of Block Size}
Figure~\ref{fig:combined_blocksize_analysis} summarizes the influence of block size on downstream task performance.

For image classification (Figure~\ref{fig:combined_blocksize_analysis}a), relatively small block sizes consistently achieve the best performance. Both the Breast and MData datasets obtain their highest F$_1$ scores using $2\times2$ blocks, although the performance differences across block sizes remain small (typically within 0.01--0.02 F$_1$). This suggests that classification performance is relatively insensitive to the precise choice of block size once the overall image structure is preserved.

The segmentation results (Figure~\ref{fig:combined_blocksize_analysis}b) exhibit a different trend. Performance is generally highest at intermediate block sizes (approximately $8\times8$), whereas both very small and very large blocks reduce segmentation quality. This observation indicates that semantic segmentation is considerably more sensitive to the choice of block size than image classification. From a practical perspective, moderate block sizes provide the best trade-off between image obfuscation and segmentation accuracy for the evaluated datasets.

\subsubsection{Effect of Noise}
Figure~\ref{fig:noise_and_blocksize_profiles} shows the effect of the optional noise component used in RMT.

Overall, model performance remains remarkably stable over a wide range of noise levels. Across both classification and segmentation tasks, changes in F$_1$ and Dice are typically within 0.01--0.02, indicating that moderate noise injection has little influence on downstream utility.

Only at the highest evaluated noise level ($256$) do we observe a noticeable decrease in classification performance. In contrast, segmentation performance remains largely unchanged and even exhibits a slight improvement on some datasets. Although the underlying mechanism requires further investigation, these results suggest that RMT is relatively insensitive to the precise choice of noise level over a broad operating range.

Overall, these studies suggest that RMT is robust to moderate changes in its transformation parameters. Small block sizes are preferable for image classification, whereas intermediate block sizes provide the best segmentation performance. In contrast, the choice of noise level has relatively little influence on downstream utility. These observations simplify practical deployment because extensive parameter tuning is generally unnecessary.

\definecolor{breastColor}{HTML}{ff7f0e}  
\definecolor{mdataColor}{HTML}{1f77b4}   
\definecolor{cvcColor}{HTML}{e74c3c}     
\definecolor{woundColor}{HTML}{1f77b4}   

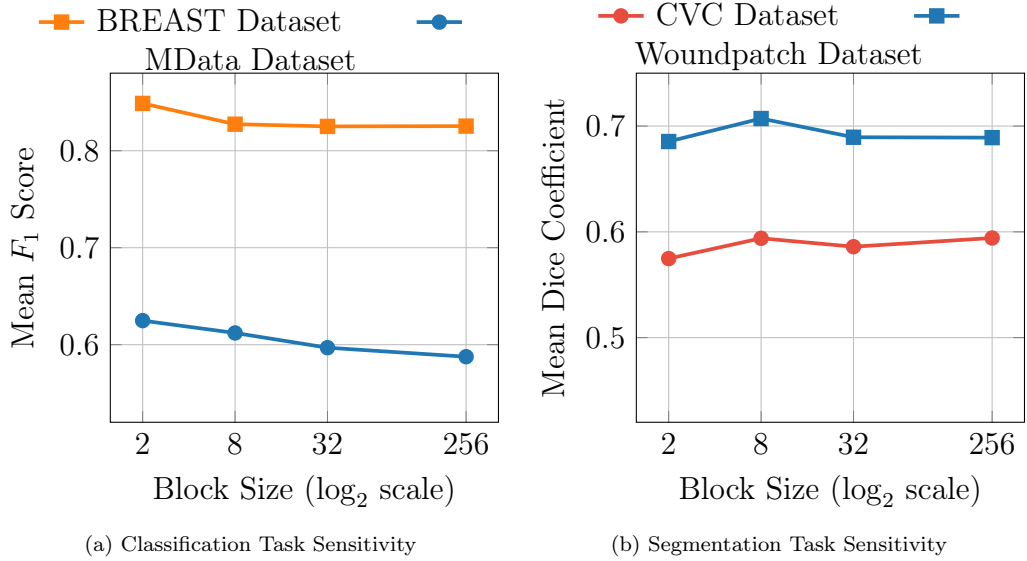
\begin{figure*}[htbp]
    \centering
    
    \begin{subfigure}[b]{0.49\textwidth}
        \centering
        \ref{lbl:breast} BREAST Dataset \qquad \ref{lbl:mdata} MData Dataset
        
        \begin{tikzpicture}
        \begin{axis}[
            xmode=log,
            log basis x=2,
            xlabel={Block Size ($\log_2$ scale)},
            ylabel={Mean $F_1$ Score},
            xtick={2,8,32,256},
            xticklabels={2,8,32,256},
            ymin=0.52, ymax=0.88, 
            grid=both,
            minor x tick num=0,
            width=\textwidth,
            height=6.2cm,
        ]
        
        \addplot[
            color=breastColor,
            mark=square*,
            line width=1.4pt,
            mark size=2.2pt
        ] coordinates {
            (2, 0.8488) 
            (8, 0.8274) 
            (32, 0.8251) 
            (256, 0.8254)
        }; \label{lbl:breast}
        
        \addplot[
            color=mdataColor,
            mark=*,
            line width=1.4pt,
            mark size=2.2pt
        ] coordinates {
            (2, 0.6248) 
            (8, 0.6121) 
            (32, 0.5969) 
            (256, 0.5876)
        }; \label{lbl:mdata}
        
        \end{axis}
        \end{tikzpicture}
        \caption{Classification Task Sensitivity}
        \label{fig:rmt-blocksize-classification}
    \end{subfigure}
    \hfill
    \begin{subfigure}[b]{0.49\textwidth}
        \centering
        \ref{lbl:cvc} CVC Dataset \qquad \ref{lbl:wound} Woundpatch Dataset
        
        \begin{tikzpicture}
        \begin{axis}[
            xmode=log,
            log basis x=2,
            xlabel={Block Size ($\log_2$ scale)},
            ylabel={Mean Dice Coefficient},
            xtick={2,8,32,256},
            xticklabels={2,8,32,256},
            ymin=0.42, ymax=0.75,
            grid=both,
            minor x tick num=0,
            width=\textwidth,
            height=6.2cm,
        ]
        
        \addplot[
            color=cvcColor,
            mark=*,
            line width=1.4pt,
            mark size=2.2pt
        ] coordinates {
            (2, 0.57474) 
            (8, 0.59392) 
            (32, 0.58596)
            (256, 0.59418)
        }; \label{lbl:cvc}
        
        \addplot[
            color=woundColor,
            mark=square*,
            line width=1.4pt,
            mark size=2.2pt
        ] coordinates {
            (2, 0.68538) 
            (8, 0.70712) 
            (32, 0.68942)
            (256, 0.68906)
        }; \label{lbl:wound}
        
        \end{axis}
        \end{tikzpicture}
        \caption{Segmentation Task Sensitivity}
        \label{fig:rmt-blocksize-segmentation}
    \end{subfigure}

    \caption{Comparative analysis of Random Matrix Transformation (RMT) block size effects on classification performance (Mean $F_1$ score) and dense segmentation performance (Mean Dice coefficient) across deep learning tasks.}
    \label{fig:combined_blocksize_analysis}
\end{figure*}

\definecolor{breastNoise}{HTML}{8e44ad}  
\definecolor{mdataNoise}{HTML}{16a085}   
\definecolor{cvcColor}{HTML}{0000FF}     
\definecolor{woundColor}{HTML}{FF0000}   

\begin{figure*}[htbp]
    \centering
    
    \begin{subfigure}[b]{0.49\textwidth}
        \centering
        \ref{lbl:breastNoise} BREAST Dataset \qquad \ref{lbl:mdataNoise} MData Dataset
        
        \begin{tikzpicture}
        \begin{axis}[
            xmode=log,
            log basis x=2,
            xlabel={Noise Level ($\log_2$ scale)},
            ylabel={Mean $F_1$ Score},
            xtick={2,8,32,256},
            xticklabels={2,8,32,256},
            ymin=0.52, ymax=0.88,
            grid=both,
            minor x tick num=0,
            width=\textwidth,
            height=6.2cm,
        ]
        
        \addplot[
            color=breastNoise,
            mark=triangle*,
            line width=1.4pt,
            mark size=2.5pt
        ] coordinates {
            (2, 0.8385) (8, 0.8540) (32, 0.8479) (256, 0.8259)
        }; \label{lbl:breastNoise}
        
        \addplot[
            color=mdataNoise,
            mark=diamond*,
            line width=1.4pt,
            mark size=2.8pt
        ] coordinates {
            (2, 0.6206) (8, 0.5899) (32, 0.6296) (256, 0.6192)
        }; \label{lbl:mdataNoise}
        
        \end{axis}
        \end{tikzpicture}
        \caption{Classification Noise Sensitivity}
        \label{fig:combined_noise_effect}
    \end{subfigure}
    \hfill
    \begin{subfigure}[b]{0.49\textwidth}
        \centering
        \ref{lbl:cvcSeg} CVC Dataset \qquad \ref{lbl:woundSeg} Woundpatch Dataset
        
        \begin{tikzpicture}
        \begin{axis}[
            xmode=log,
            log basis x=2,
            xlabel={Noise Level ($\log_2$ scale)},
            ylabel={Segmentation Metric DICE},
            xmin=1.5, xmax=300,
            ymin=0.55, ymax=0.72,
            xtick={2, 8, 32, 256},
            xticklabels={2,8,32,256},
            grid=both,
            width=\textwidth,
            height=6.2cm,
        ]
        
        \addplot[
            color=cvcColor,
            mark=square*,
            thick
        ] coordinates {
            (2,0.57474) 
            (8,0.59392) 
            (32,0.58596) 
            (256,0.59418)
        }; \label{lbl:cvcSeg}
        
        \addplot[
            color=woundColor,
            mark=*,
            thick
        ] coordinates {
            (2,0.68192) 
            (8,0.69264) 
            (32,0.68732) 
            (256,0.68906)
        }; \label{lbl:woundSeg}
        
        \end{axis}
        \end{tikzpicture}
        \caption{Segmentation Noise Sensitivity}
        \label{fig:rmt_noise_seg}
    \end{subfigure}

    \caption{Downstream task parameters under RMT: comparative impact of varying noise dimensions on evaluation metrics across diagnostic frameworks.}
    \label{fig:noise_and_blocksize_profiles}
\end{figure*}
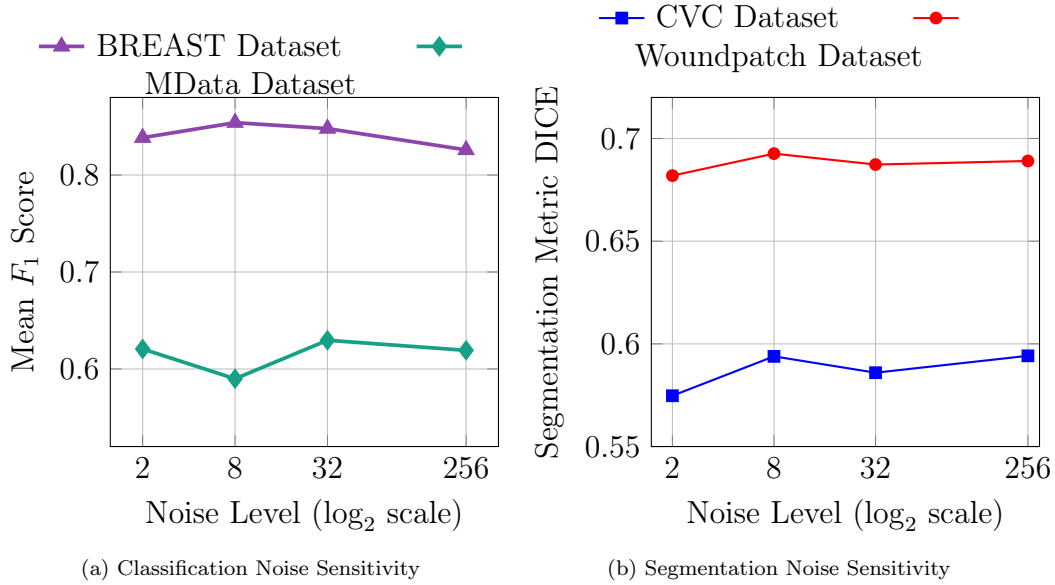

\subsubsection{Resilience to Known-Pair Regression Attacks}
The security of RMT relies on the confidentiality of the block-wise transformation matrices $\{R_b\}$. To evaluate its robustness under a worst-case information leakage scenario, we consider the regression attack proposed in the original DisguisedNets work \cite{chen23toit}. The attacker is assumed to possess a number of original image--disguised image pairs and attempts to estimate the underlying block-wise transformation matrices through linear regression. Once the estimated matrices are obtained, they are used to reconstruct previously unseen disguised images.

Previous studies demonstrated that this attack can partially recover images from low-resolution natural-image datasets such as MNIST and CIFAR-10. Whether the same attack remains effective for realistic medical images has not previously been investigated.

To evaluate attack effectiveness, we consider the most favorable setting for the attacker by using the smallest block size ($2\times2$), where the regression model is expected to estimate the transformation most accurately. We further investigate whether the optional noise component in RMT provides additional protection. Attack performance is evaluated quantitatively using the DNN examiner introduced in Section~\ref{sec:method} and qualitatively through visual inspection of the reconstructed images.

For the Breast classification task (Figure~\ref{fig:global_known_pair_resilience}a), the reconstructed images remain largely unrecognizable to the pretrained DNN examiner. When only a small number of image pairs are available ($K\le3$), the examiner performance remains close to random guessing ($F_1\approx0.50$). Surprisingly, increasing the number of leaked image pairs does not improve the attack. Instead, the examiner performance gradually decreases, reaching approximately $F_1=0.45$ when $K=100$. This observation suggests that the regression estimator does not converge toward a more accurate inverse transformation on these medical images. Rather, the reconstructed images become progressively less recognizable to the downstream classifier. One possible explanation is that estimation errors accumulate systematically as the regression model attempts to fit the high-dimensional transformation, resulting in reconstructed images that deviate further from the true image manifold. Further investigation is needed to fully understand this behavior.

The segmentation results (Figure~\ref{fig:global_known_pair_resilience}b) demonstrate even stronger resistance. Across all leakage levels and both noiseless and noisy configurations, the reconstructed images produce Dice scores close to zero, indicating that the regression attack fails to recover spatial information sufficient for meaningful semantic segmentation.

\begin{figure*}[htbp]
    \centering
    \includegraphics[width=\textwidth]{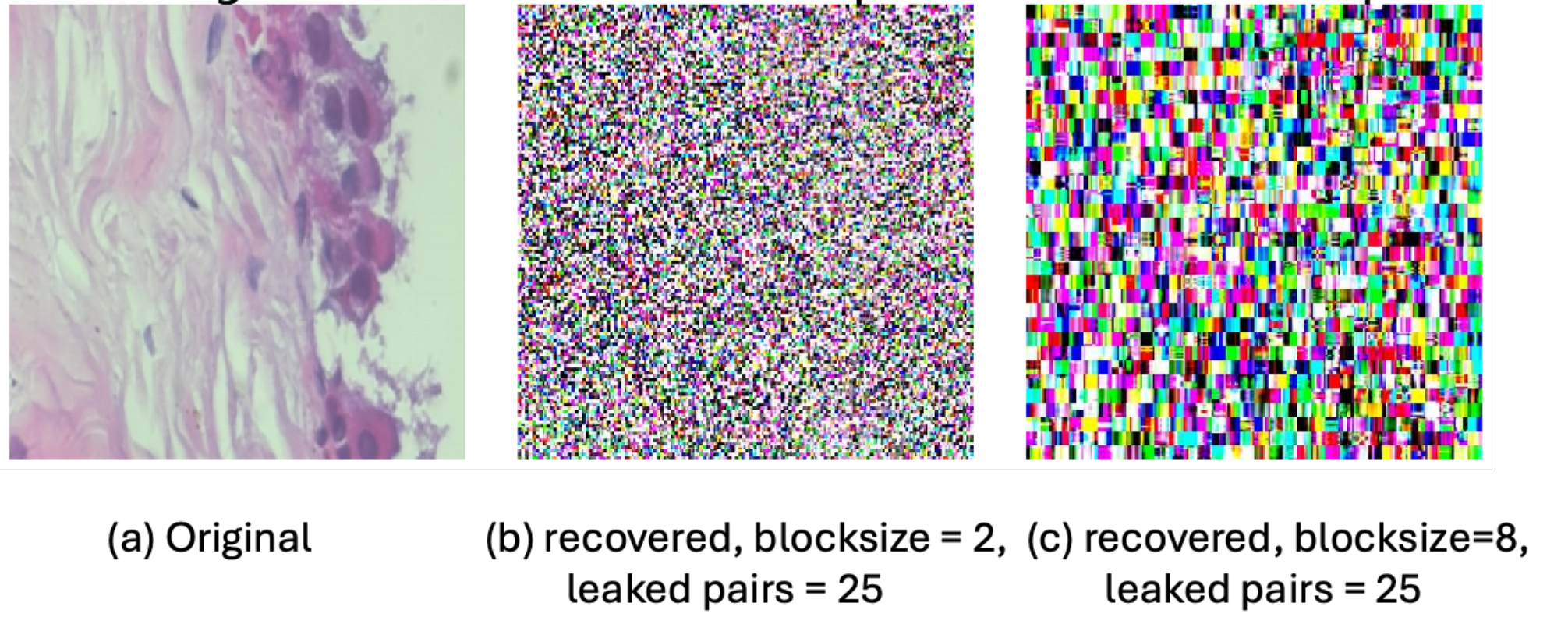} 
    \caption{Visual comparison of known-pair regression attack reconstruction fidelity at $K=25$ leaked pairs for a Breast Cancer sample image: (a) unmodified original image, (b) reconstructed image from an RMT configuration utilizing a small block size ($b=2$), and (c) reconstructed image from a larger block size profile ($b=8$). Clearly no visual clues are recovered.}
    \label{fig:breast_attack_reconstruction_comparison}
\end{figure*}

\begin{figure*}[htbp]
    \centering
    \includegraphics[width=\textwidth]{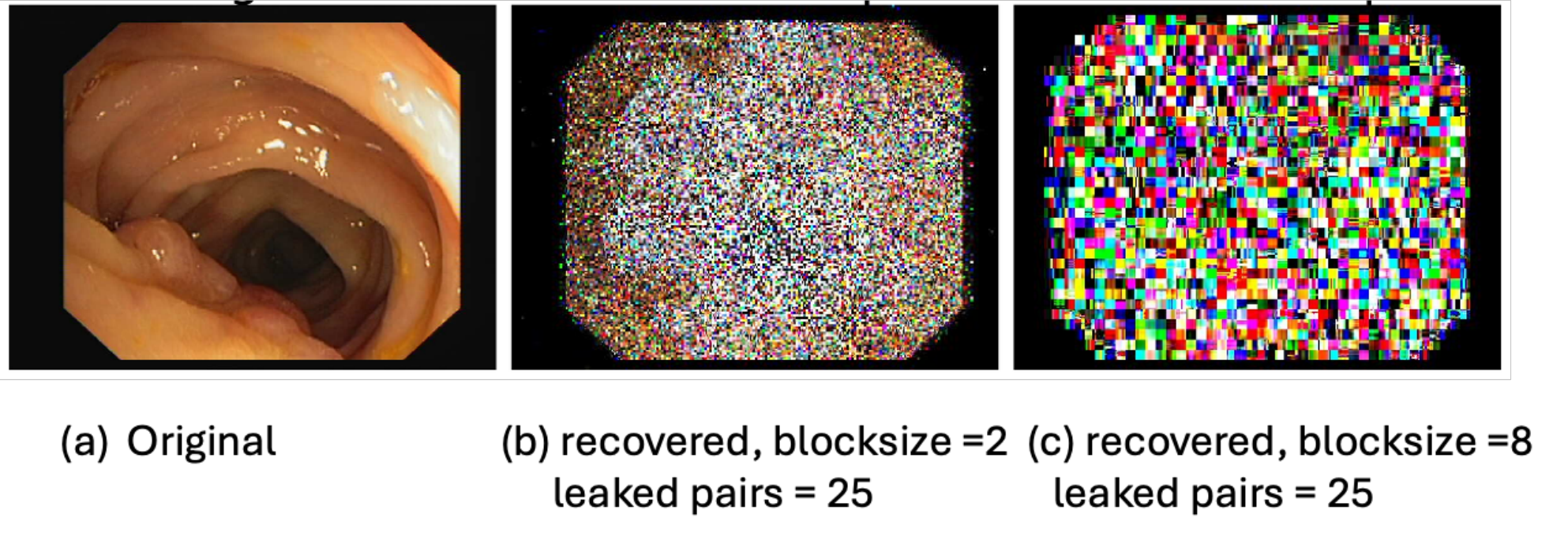} 
    \caption{Visual comparison of known-pair regression attack reconstruction fidelity at $K=25$ leaked pairs for a CVC sample image: (a) unmodified original image, (b) reconstructed image from an RMT configuration utilizing a small block size ($b=2$), and (c) reconstructed image from a larger block size profile ($b=8$). Clearly no visual clues are recovered.}
    \label{fig:cvc_attack_reconstruction_comparison}
\end{figure*}

The quantitative results are consistent with the visual reconstruction examples shown in Figures~\ref{fig:breast_attack_reconstruction_comparison} and \ref{fig:cvc_attack_reconstruction_comparison}. Even after observing 25 original image--disguised image pairs, the reconstructed Breast and CVC images reveal little recognizable anatomical structure under either fine ($2\times2$) or moderate ($8\times8$) block sizes. The reconstructed images remain visually distorted and do not expose clinically meaningful information.

Overall, these results suggest that the regression attack, although previously shown to be partially effective on low-resolution natural-image datasets, is considerably less effective on the medical imaging datasets evaluated in this study. The increased image resolution and structural complexity appear to make accurate estimation of the transformation matrices substantially more difficult, thereby providing stronger practical privacy protection under the evaluated attack model.

\definecolor{noiseZeroColor}{HTML}{d35400}  
\definecolor{noiseEightColor}{HTML}{2980b9} 

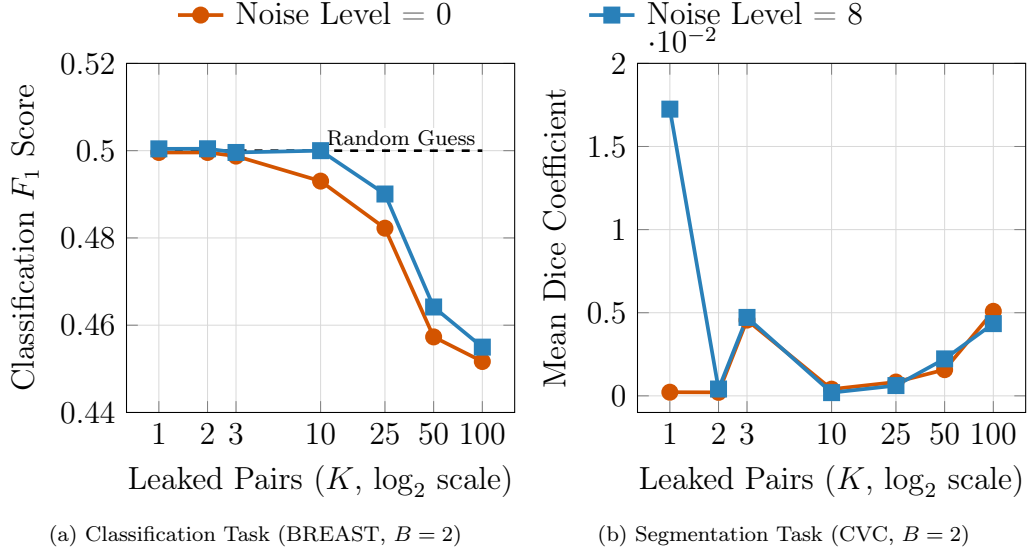
\begin{figure*}[htbp]
    \centering
    
    \ref{lbl:noiseZero} Noise Level = 0 \qquad\qquad \ref{lbl:noiseEight} Noise Level = 8

    \begin{subfigure}[b]{0.49\textwidth}
        \centering
        \begin{tikzpicture}
        \begin{axis}[
            xmode=log,
            log basis x=2,
            xlabel={Leaked Pairs ($K$, $\log_2$ scale)},
            ylabel={Classification $F_1$ Score},
            xtick={1, 2, 3, 10, 25, 50, 100},
            xticklabels={1, 2, 3, 10, 25, 50, 100},
            ymin=0.44, ymax=0.52,
            grid=both,
            grid style={line width=.1pt, draw=gray!10},
            major grid style={line width=.2pt, draw=gray!30},
            minor x tick num=0,
            width=\textwidth,
            height=6.2cm,
        ]

\addplot[
    black,
    dashed,
    line width=1pt
] coordinates {
    (1,0.50)
    (100,0.50)
};

\node[
    anchor=south east,
    font=\scriptsize,
    fill=white,
    inner sep=1pt
] at (axis cs:95,0.5005) {Random Guess};

        \addplot[
            color=noiseZeroColor,
            mark=*,
            line width=1.4pt,
            mark size=2.5pt
        ] coordinates {
            (1, 0.499585) (2, 0.499587) (3, 0.498755) (10, 0.493021)
            (25, 0.482232) (50, 0.457318) (100, 0.451683)
        }; \label{lbl:noiseZero} 

        \addplot[
            color=noiseEightColor,
            mark=square*,
            line width=1.4pt,
            mark size=2.5pt
        ] coordinates {
            (1, 0.500415) (2, 0.500434) (3, 0.499585) (10, 0.500000)
            (25, 0.490066) (50, 0.464173) (100, 0.454977)
        }; \label{lbl:noiseEight} 

        \end{axis}
        \end{tikzpicture}
        \caption{Classification Task (BREAST, $B=2$)}
        \label{fig:rmt_known_pair_attack_breast}
    \end{subfigure}
    \hfill
    \begin{subfigure}[b]{0.49\textwidth}
        \centering
        \begin{tikzpicture}
        \begin{axis}[
            xmode=log,
            log basis x=2,
            xlabel={Leaked Pairs ($K$, $\log_2$ scale)},
            ylabel={Mean Dice Coefficient},
            xtick={1, 2, 3, 10, 25, 50, 100},
            xticklabels={1, 2, 3, 10, 25, 50, 100},
            ymin=-0.001, ymax=0.020,
            grid=both,
            grid style={line width=.1pt, draw=gray!10},
            major grid style={line width=.2pt, draw=gray!30},
            minor x tick num=0,
            width=\textwidth,
            height=6.2cm,
        ]

        \addplot[
            color=noiseZeroColor,
            mark=*,
            line width=1.4pt,
            mark size=2.5pt
        ] coordinates {
            (1, 0.000219) (2, 0.000210) (3, 0.004558) (10, 0.000401)
            (25, 0.000830) (50, 0.001575) (100, 0.005085)
        };

        \addplot[
            color=noiseEightColor,
            mark=square*,
            line width=1.4pt,
            mark size=2.5pt
        ] coordinates {
            (1, 0.017247) (2, 0.000411) (3, 0.004718) (10, 0.000194)
            (25, 0.000618) (50, 0.002223) (100, 0.004345)
        };

        \end{axis}
        \end{tikzpicture}
        \caption{Segmentation Task (CVC, $B=2$)}
        \label{fig:rmt_known_pair_attack_cvc}
    \end{subfigure}

    \caption{Resilience comparison of RMT image disguising against known-pair regression attacks under varying noise configurations across classification and segmentation architectures. F1=0.5 for binary classification is equivalent to random guesses.}
    \label{fig:global_known_pair_resilience}
\end{figure*}

\section{Discussion}\label{sec:discussion}

This study provides the first systematic evaluation of representative image disguising methods for confidential medical image analysis. Although image disguising has demonstrated encouraging results on natural-image benchmarks, our experiments show that these conclusions do not directly generalize to clinically relevant medical imaging tasks. More importantly, the observed performance differences reveal several fundamental characteristics that should guide the future design of privacy-enhancing technologies (PETs) for confidential medical AI outsourcing.

\subsection{Different Medical AI Tasks Respond Differently to Disguised Images}

The most important finding of this study is that image classification and semantic segmentation respond fundamentally differently to image disguising. 
Current disguising techniques appear suitable for cloud-based disease screening, triage, and image-level diagnostic support, where image-level predictions are sufficient. However, they are currently inadequate for image-guided interventions, lesion contouring, radiation therapy planning, and other applications requiring accurate spatial localization.

A likely explanation is that image-level classification primarily depends on learning discriminative semantic representations, whereas semantic segmentation additionally requires accurate preservation of local spatial relationships. Block-wise image transformations inevitably disturb these spatial relationships even when sufficient global semantic information remains for classification. Consequently, preserving visual confidentiality alone is insufficient for dense medical image analysis. Future privacy-enhancing techniques should explicitly preserve clinically meaningful spatial structures while maintaining visual obfuscation.

These observations also suggest that evaluating privacy-preserving methods solely on natural-image classification benchmarks may substantially overestimate their applicability to medical AI. Clinically relevant evaluation should include dense prediction tasks that better reflect real-world diagnostic workflows.

\subsection{Implications for Confidential Medical AI}
From a deployment perspective, our results indicate that image disguising is already a practical solution for several classes of medical AI applications. Among the evaluated methods, RMT consistently achieved the best overall balance among predictive utility, computational efficiency, and robustness against reconstruction attacks. Moreover, its performance remained relatively stable across different block sizes and noise levels, suggesting that extensive parameter tuning may not be necessary in practice.

The observed task-dependent behavior also provides practical guidance for selecting privacy-preserving strategies. Current image disguising methods appear well suited for image-level applications such as disease screening, image triage, and diagnostic classification, where global semantic information dominates model predictions. In contrast, applications requiring precise spatial localization, including tumor delineation, organ segmentation, and wound boundary estimation, remain challenging because current disguising methods disrupt the fine anatomical structures required for dense prediction.

\subsection{Balancing Privacy, Utility, and Computational Efficiency}

An important motivation for image disguising is to bridge the gap between conventional encryption and practical deep learning. Compared with cryptographic approaches such as homomorphic encryption and secure multi-party computation, image disguising introduces only a lightweight preprocessing step while allowing existing deep learning pipelines to operate without modification. Our experiments confirm that preprocessing requires only milliseconds per image and does not noticeably increase model convergence time, making the computational cost negligible compared with model training itself.

The security evaluation further demonstrates that reconstruction attacks become considerably less effective on realistic medical images than previously reported on low-resolution natural-image datasets. Even under the strongest evaluated known-pair attack, reconstructed images remained visually unintelligible and yielded little useful information for downstream classification or segmentation. Interestingly, increasing the number of leaked image pairs did not improve attack performance and, for image classification, even reduced performance below the random-guessing baseline. Although the underlying mechanism deserves further theoretical investigation, these results suggest that the structural complexity of medical images substantially increases the difficulty of regression-based reconstruction.

Taken together, these findings indicate that image disguising represents a practical point in the privacy-utility-efficiency design space. While it does not provide the rigorous security guarantees of cryptographic computation, it offers substantially lower computational cost while preserving useful predictive performance for many clinically relevant image-level tasks.

\subsection{Limitations and Future Directions}
This study has several limitations. First, although four representative datasets were evaluated, additional imaging modalities including CT, MRI, retinal imaging, and ultrasound should be investigated to further validate the generality of our conclusions. Second, we focused on two representative image disguising frameworks. Emerging privacy-enhancing image transformation methods may exhibit different utility--privacy trade-offs and should be evaluated using the proposed framework. Third, our security analysis concentrated on regression-based reconstruction attacks following the original DisguisedNets threat model. Future work should investigate stronger adversaries based on modern generative and diffusion models. Finally, while our experiments consistently show that realistic medical images are substantially more resistant to regression-based reconstruction than natural-image benchmarks, developing a theoretical explanation for this phenomenon remains an important direction for future research.

Overall, our findings suggest that image disguising is already a promising privacy-enhancing technology for confidential medical AI, particularly for cloud-based image classification and screening applications. At the same time, the substantial utility degradation observed in dense segmentation highlights an important limitation of current approaches and points toward a key research direction for future privacy-preserving medical image learning: jointly preserving patient confidentiality and clinically meaningful spatial structure.

\section{Conclusion}
\label{sec:conclusion}
This study establishes the first systematic evaluation of representative image disguising methods for confidential medical image analysis. Using a unified experimental framework spanning four public medical imaging datasets, we compared representative image disguising approaches across image classification and semantic segmentation with respect to predictive utility, computational efficiency, parameter sensitivity, and robustness against reconstruction attacks.

Our results demonstrate that image disguising represents a practical privacy-enhancing technology for cloud-based medical AI. Compared with conventional cryptographic approaches, it introduces only lightweight preprocessing overhead while remaining compatible with standard deep learning pipelines. Among the evaluated methods, Randomized Multidimensional Transformation (RMT) consistently achieved the best overall balance between predictive performance, computational efficiency, and resistance to reconstruction attacks, whereas AES-based disguising incurred substantial utility degradation.

More importantly, this study reveals a fundamental distinction between image-level and pixel-level medical AI tasks. While current image disguising methods preserve sufficient semantic information for medical image classification, they fail to adequately preserve the fine-grained spatial structure required for dense semantic segmentation. These findings suggest that preserving visual confidentiality alone is insufficient for confidential medical image analysis and identify spatial structure preservation as a key design requirement for future privacy-enhancing technologies.

Overall, this work provides the first systematic evidence regarding the capabilities and limitations of image disguising for medical AI and establishes a reproducible evaluation framework for future research. We hope these findings will facilitate the development of next-generation privacy-enhancing technologies that jointly optimize privacy, predictive utility, and computational efficiency, enabling trustworthy and scalable cloud-based medical AI systems.

\section*{Funding}
This material is based upon work partially supported by the National Science Foundation under Grant No. 2517121. Any opinions, findings, and conclusions or recommendations expressed in this material are those of the author(s) and do not necessarily reflect the views of the National Science Foundation.

\section*{Ethics Statement}
This study used publicly available, de-identified datasets and did not involve interaction with human participants. Institutional review board approval and informed consent were therefore not required.

\section*{Declaration of competing interest}
The authors declare that they have no known competing financial interests or personal relationships that could have appeared to influence the work reported in this paper.

\section*{Declaration of generative AI and AI-assisted technologies in the manuscript preparation process}
During the preparation of this work, the authors used ChatGPT to assist with language editing and manuscript organization. After using this tool, the authors reviewed and edited the content as needed and take full responsibility for the content of the published article.

\section*{CRediT authorship contribution statement}

Keke Chen: Conceptualization, Methodology, Supervision,
Funding acquisition, Formal analysis,
Writing – review \& editing.

Jason Rojas, Jiajie He, and Yash Patel: Software, Data curation,
Investigation, Validation,
Visualization, Writing – original draft.

Yuechun Gu: Software, Data curation, Investigation, Validation

Zeyun Yu: Data curation, Supervision, Writing – review \& editing.

\section*{Data Availability}

All datasets used in this study are publicly available from their original sources. Links to the datasets and the complete evaluation framework are provided in the manuscript and the accompanying GitHub repository.

\section*{Code Availability}
The complete evaluation framework is publicly available at
\url{https://github.com/Jasonmix84/Secure-By-Disguise/}.

\bibliographystyle{elsarticle-num}
\bibliography{paper,dl,health}

\end{document}